\pdfoutput=1

\documentclass[11pt]{article}

\usepackage{ACL2023}

\usepackage{times}
\usepackage{latexsym}

\usepackage[T1]{fontenc}

\usepackage[utf8]{inputenc}

\usepackage{microtype}

\usepackage{inconsolata}
\usepackage{graphicx}
\usepackage{float}

\usepackage{amsmath}
\usepackage{amssymb}
\usepackage{hyperref}
\usepackage{caption}
\usepackage{subcaption}

\usepackage{natbib}
\usepackage[nameinlink]{cleveref}

\title{Anisotropy Is Inherent to Self-Attention in Transformers}

\author{
    Nathan Godey$\mkern6mu^{1,2}$ \quad Éric de la Clergerie$^1$ \quad Benoît Sagot$^1$ \\
    $^1$Inria, Paris, France \\
    $^2$Sorbonne Université, Paris, France \\
    \texttt{\{nathan.godey,eric.de\_la\_clergerie,benoit.sagot\}@inria.fr}
}

\begin{document}
\maketitle
\begin{abstract}
The representation degeneration problem is a phenomenon that is widely observed among self-supervised learning methods based on Transformers. In NLP, it takes the form of \textit{anisotropy}, a singular property of hidden representations which makes them unexpectedly close to each other in terms of angular distance (cosine-similarity). Some recent works tend to show that anisotropy is a consequence of optimizing the cross-entropy loss on long-tailed distributions of tokens. We show in this paper that anisotropy can also be observed empirically in language models with specific objectives that should not suffer directly from the same consequences. We also show that the anisotropy problem extends to Transformers trained on other modalities. Our observations suggest that anisotropy is actually inherent to Transformers-based models.
\end{abstract}

\section{Introduction}
In recent years, deep learning models based on Transformers have led to significant breakthroughs in the field of natural language processing (NLP). These models have demonstrated state-of-the-art performance across a range of tasks, such as language modeling, machine translation, and sentiment analysis. However, despite their successes, they suffer from a phenomenon known as the representation degeneration problem. Specifically, this degeneration is characterized by anisotropy, a property of hidden representations that makes them all close to each other in terms of angular distance (cosine-similarity).

Anisotropy has been widely observed among self-supervised models based on Transformers, and literature currently suggests that it may be a consequence of optimizing the cross-entropy loss on long-tailed distributions of tokens \citep{GaoHTQWL19, bis-etal-2021-much}. However, it remains uncertain whether anisotropy is a fundamental property of Transformers-based models or a consequence of the pre-training process.

In this paper, we investigate the anisotropy problem in depth, and we make several contributions:
\begin{itemize}
    \item We demonstrate empirically that anisotropy can be observed in language models with character-aware architectures that should not suffer directly from the same consequences as token-based models. We extend our observations to Transformers trained on other modalities, such as image and audio data, and show that anisotropy cannot be explained solely based on linguistic properties;
    \item We provide empirical observations on the anisotropic properties of the Transformer block by studying untrained layers, and establish a relation between anisotropy and the general sharpness of the self-attention mechanism;
    \item We conduct an analysis of the representations used in self-attention (queries and keys) along training and show that anisotropy appears intrinsically in the self-attention mechanism, when training pushes for sharp patterns.
\end{itemize} 

\section{Related Work}

\begin{figure}[h]
    \centering
     \includegraphics[width=0.9\columnwidth]{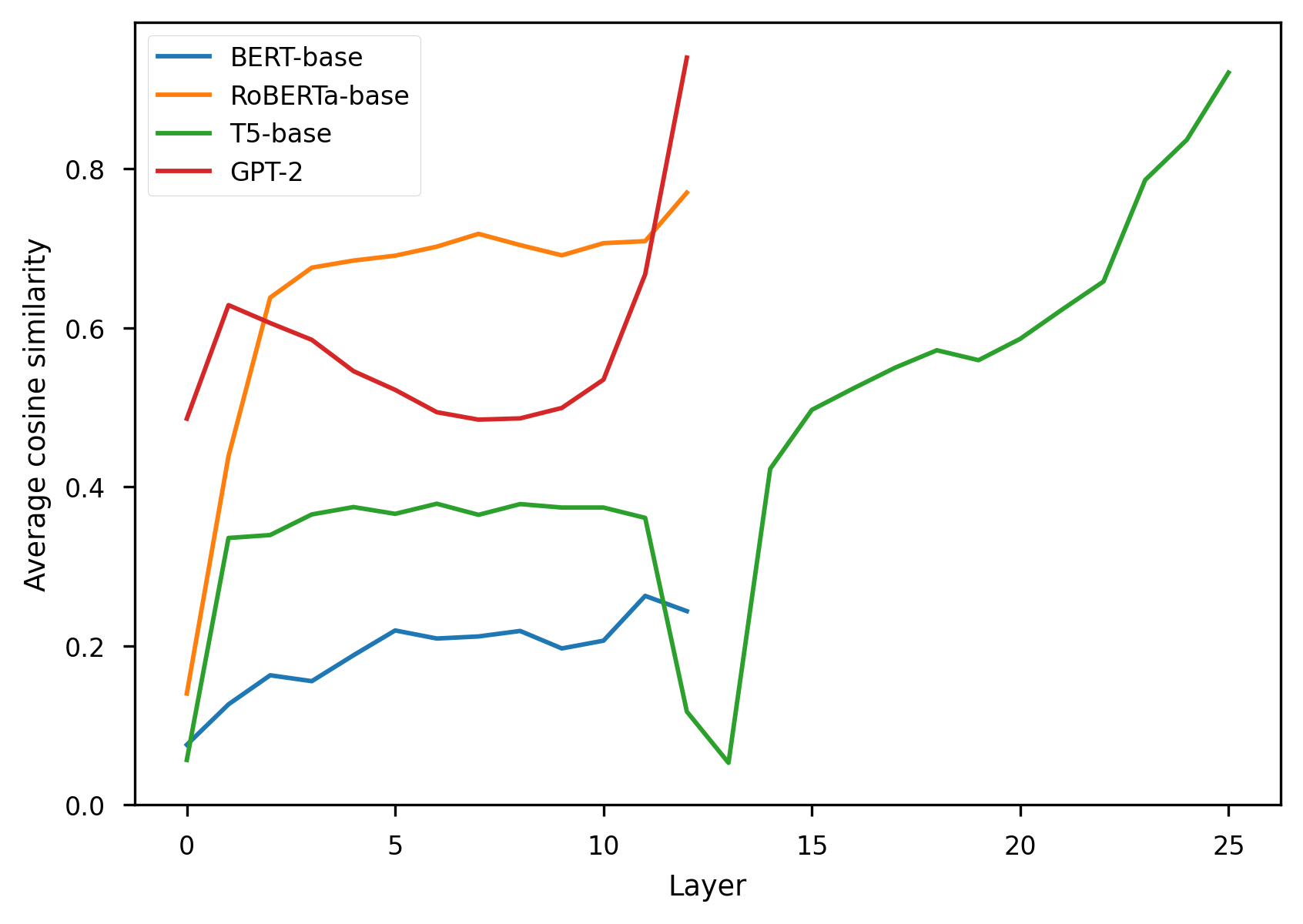}
     \caption{Average cosine-similarity between hidden representations across layers for token-level NLP models. For T5-base, we concatenate encoder and decoder results.}
     \label{fig:anisotropy_token}
\end{figure}

The general phenomenon of anisotropy in token-based Transformers for language models has been shown in \citet{ethayarajh-2019-contextual}. \autoref{fig:anisotropy_token} extends one of their experiment to more architectures. \citet{GaoHTQWL19} shows that the degeneration of representations comes from the distributions of subwords in natural language, namely the existence of unused and rare tokens that tend to push all representations away from the origin towards a specific direction.

Other works have established a connection between word frequency and distortions of the latent spaces \citep{yu-etal-2022-rare, puccetti-etal-2022-outlier, rajaee-pilehvar-2022-isotropy}. \citet{bis-etal-2021-much} have shown that anisotropy in LMs could be explained by a global \textit{drift} of the representations in the same direction, thus unifying conclusions from \citet{ethayarajh-2019-contextual} and \citet{GaoHTQWL19}. The authors propose that this drift is caused by the persistent updating of the representation of rare and unused tokens in a consistent direction, due to the nature of the softmax operation in the cross-entropy loss. They show that removing the average component to all representations leads to a nearly perfect isotropy.

Several methods have been proposed to reduce anisotropy in Transformers-based LMs at token-level \citep{rajaee-pilehvar-2021-cluster, Wang2020Improving}, or at sentence-level \citep{gao-etal-2021-simcse, yan-etal-2021-consert, su2021whitening}. They usually consist in post-processing the representations, and lead to downstream performance boosts. We argue that these positive results are paving the way for the search of pre-training objectives that do not introduce anisotropy in the first place, in the hope that the resulting models will also perform better without any post-processing, and potentially be trained more efficiently. This motivates us to gain a deeper understanding of the underlying factors that induce anisotropy, whether they belong in data, architectures, or training procedures.

\section{Anisotropy in pre-trained Transformers}
\subsection{Character-based NLP}
\label{sec:charbased}
\begin{figure}[h]
    \centering
     \includegraphics[width=0.9\columnwidth]{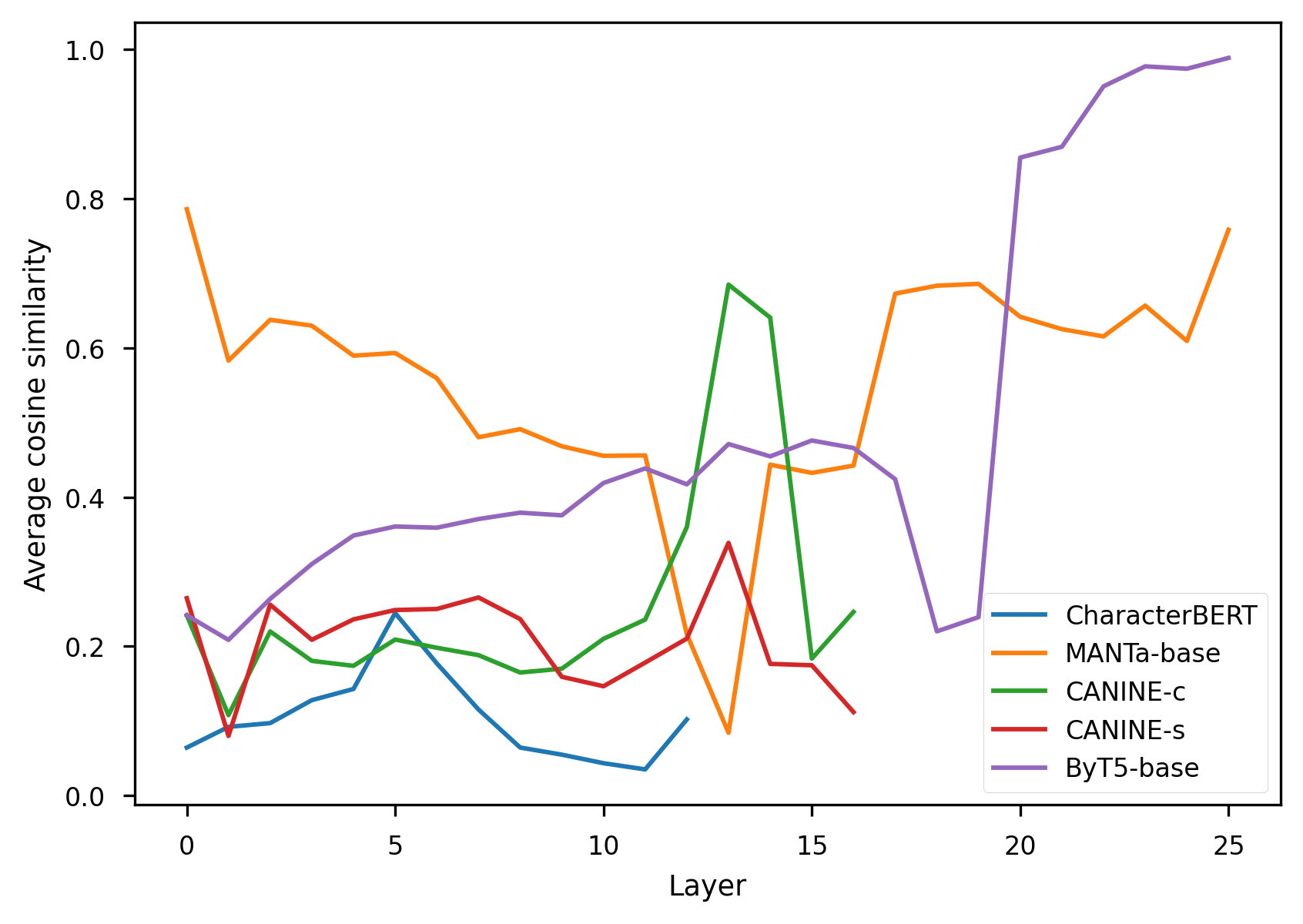}
     \caption{Average cosine-similarity between hidden representations across layers for character-level models.}
     \label{fig:cos_char_aware}
\end{figure}

To assert whether the cross-entropy objective applied on vocabularies containing rare tokens is the sole cause for the common drift issue, we explore anisotropy in character-based models. We study different architectures:
\begin{itemize}
    \item CharacterBERT \citep{el-boukkouri-etal-2020-characterbert} is constructing whole word representations from character embeddings put through convolutions and highway layers, before feeding them to a Transformers architecture.
    \item CANINE \citep{clark-etal-2022-canine} is downsampling contextualized character representations via a strided convolution before feeding them to a Transformers. It can be trained either with a subword-based objective (CANINE-s) or with a character-level one (CANINE-c).
    \item MANTa-LM \citep{godey-etal-2022-manta} is based on a differentiable segmentation and embedding module added before an encoder-decoder model in the style of T5 \citep{2020t5}. It takes bytes as inputs and outputs, but builds internal representations that are usually based on several bytes.
    \item ByT5 \citep{xue-etal-2022-byt5} is a version of T5 that is trained at byte-level. To afford for more complex encoding, the authors resize the encoder-decoder architecture.
\end{itemize}

Neither of these architectures should suffer from out-of-vocabulary tokens in the process of creating representations. The models that predict at word or sub-word level (CharacterBERT and CANINE-s) could have the cross-entropy loss systematically pushing away rare item representations. However, it is rather unclear why it would imply an embedding drift at deeper layers. Hence, if anisotropy was only caused by the presence of unused or rare subwords, those character-level models should be much less prone to this issue.

To verify this hypothesis, we compute hidden representations for the validation set of the WikiText-103 corpus \citep{MerityXBS16}. We then compute the average cosine-similarity between two representations, uniformly taken in the whole validation corpus.

In fact, as shown in \autoref{fig:cos_char_aware}, those models all display significant levels of anisotropy in at least one of their layers. Interestingly, the models that are based solely on characters or bytes for input and prediction (ByT5, CANINE-c, and MANTA-LM) seem to display even higher levels of anisotropy. We note, as it is the case for the T5 model, that the ByT5 decoder displays extremely high levels of anisotropy.

\subsection{Other modalities}
\label{sec:other_mod}
\begin{figure*}[h]
    \centering
    \begin{subfigure}[b]{0.43\textwidth}
         \includegraphics[width=\linewidth]{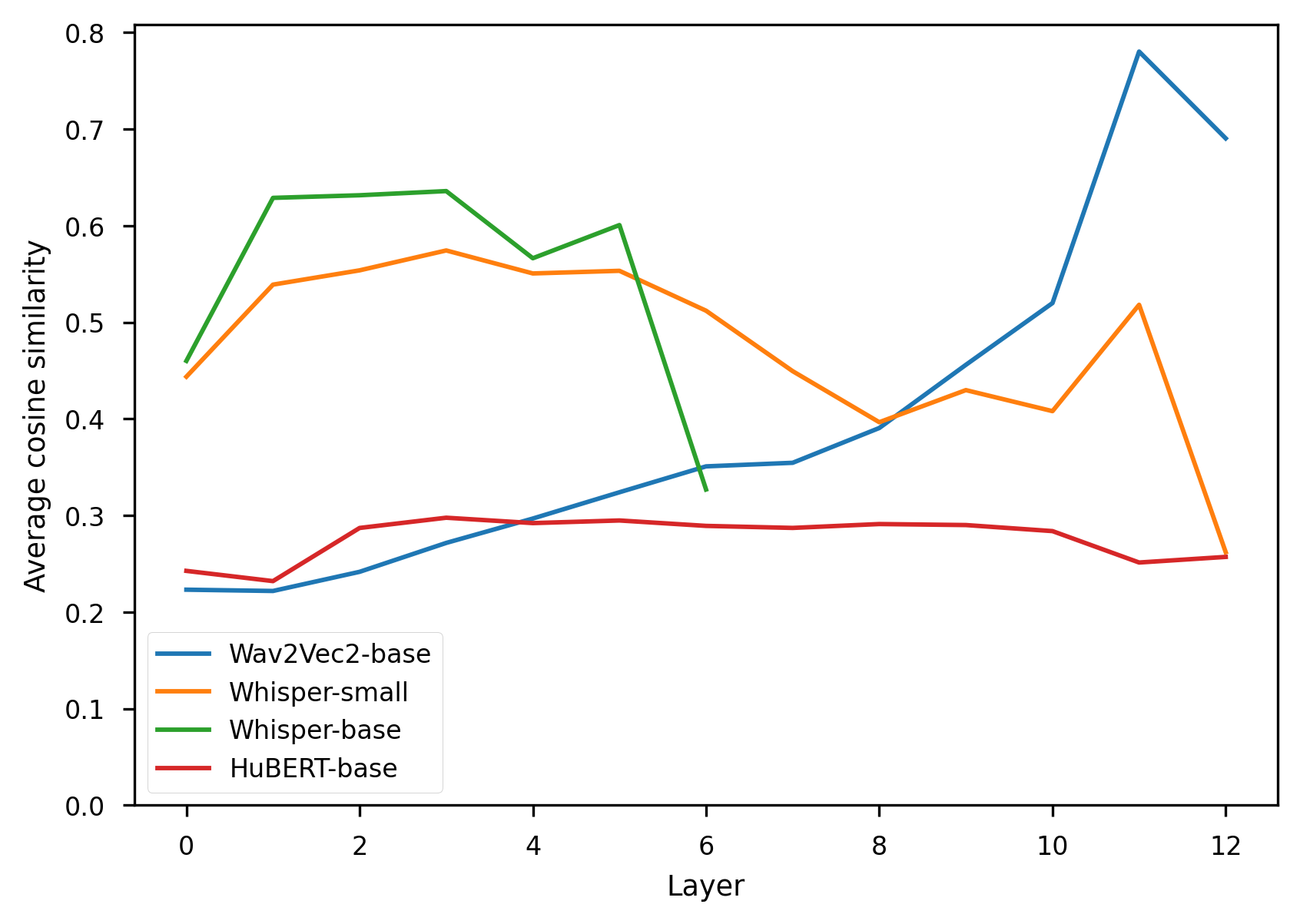}
         \caption{Speech}
         \label{fig:cos_speech}
    \end{subfigure}
    \hfill
    \begin{subfigure}[b]{0.43\textwidth}
         \includegraphics[width=\linewidth]{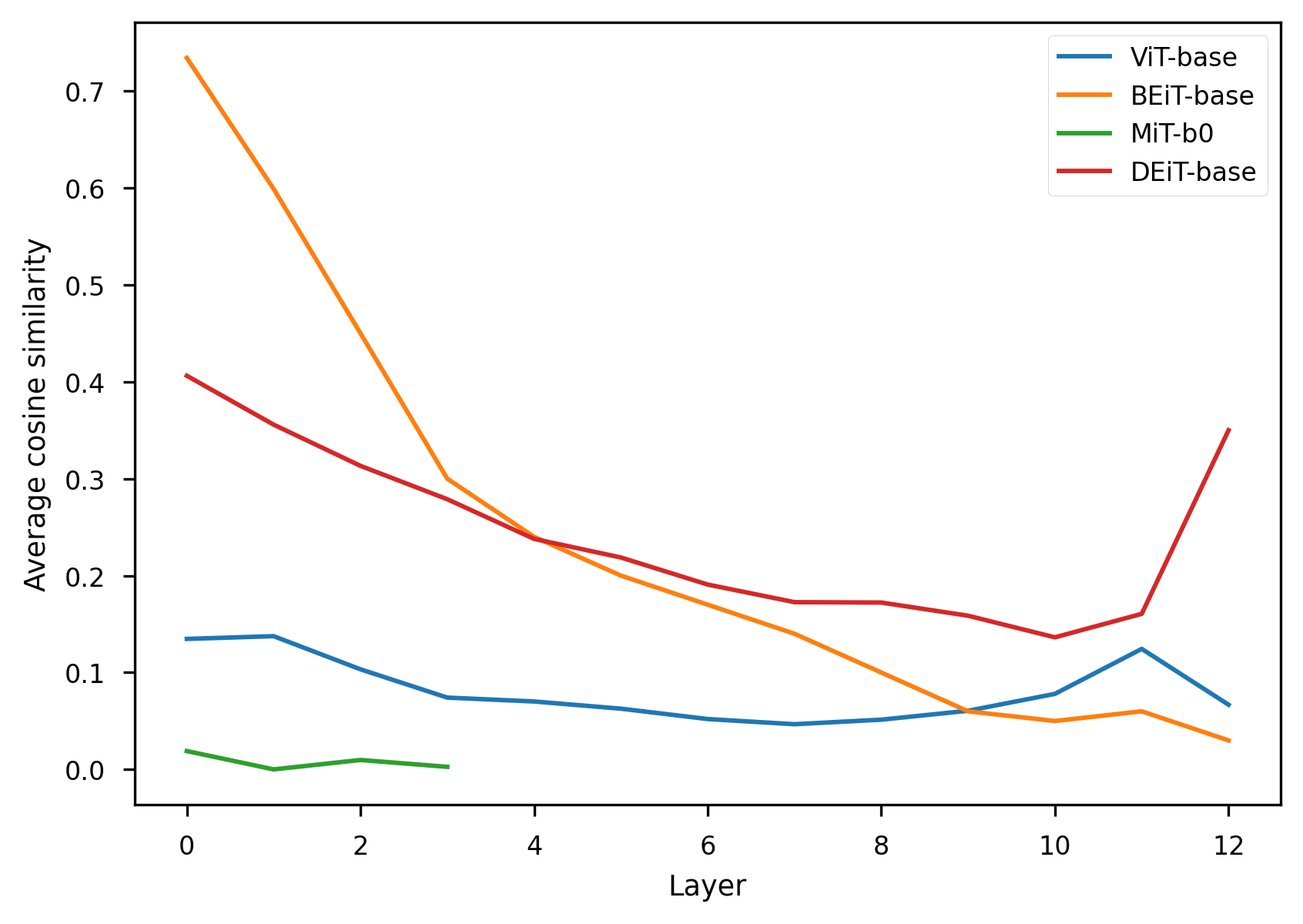}
         \caption{Vision}
         \label{fig:cos_audio}
    \end{subfigure}
    \caption{Average cosine-similarity between hidden representations across layers for Speech and Vision modalities. We observe that across both modalities, several models display significant levels of anisotropy.}
    \label{fig:anisotropy_modalities}
\end{figure*}
We've shown in the previous section that character-level language models suffer from anisotropy similarly to token-level ones, hinting that subword token distributions are not solely responsible for anisotropy. However, it may be argued that anisotropy is related to linguistic properties. Thus, we proceed to explore the anisotropy problem for Transformers-based models in other modalities, specifically speech and vision.

For speech models, we consider wav2Vec 2.0 \citep{wav2vec}, HuBERT \citep{HuBERT}, and Whisper \citep{radford2022whisper} with the Common Voice 11.0 dataset \citep{commonvoice:2020}. For vision models, we use ViT \citep{Wu2020VisualTT}, BEiT \citep{beit-2021}, MiT \citep{segformer21}, and DEiT \citep{pmlr-v139-touvron21a} on the ImageNet dataset \citep{imagenet15russakovsky}.

As in \autoref{sec:charbased}, we infer hidden representations on the validation sets for each modality. We then uniformly sample pairs of vectors to get cosine-similarity values for every layer of every model. The averaged results are displayed in \autoref{fig:anisotropy_modalities}.

Once again, almost every model shows a significant level of anisotropy on some of its layers. Notably, speech models seem to have very anisotropic representations, as every layer of every model outputs an average cosine-similarity of at least $0.2$. We find some exceptions among vision models, since the MiT model seems to use isotropic representation spaces and the ViT model has a low average cosine-similarity for all its layers.

We also conduct the same experiment for convolution-based networks in the vision modality. The models at glance are ResNet \citep{he2016deep}, EfficientNet \citep{Tan2019EfficientNetRM}, CvT \citep{wu2021cvt}, ConvNeXt \citep{liu2022convnet}, and VAN \citep{guo2022visual}. For these networks, we flatten convolution maps to vectors before computing the cosine-similarity.

\begin{figure}[H]
    \centering
    \includegraphics[width=\linewidth]{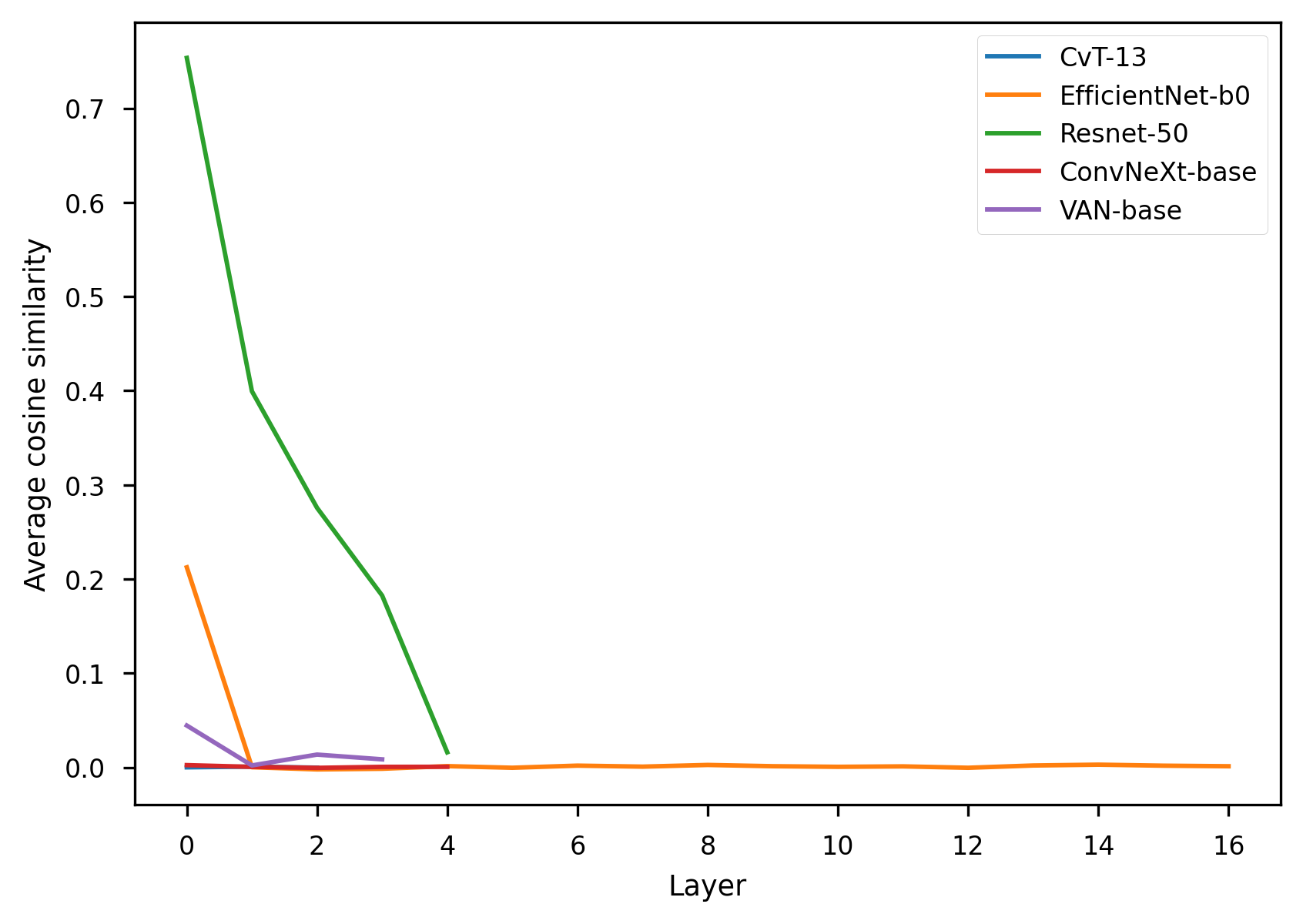}
    \caption{Average cosine-similarity between hidden representations across layers for convolution-based vision models.}
    \label{fig:convbased}
\end{figure}

We observe in \autoref{fig:convbased} that most of the convolution-based models are isotropic. Interestingly, the only exception is ResNet-50, whose representations become more and more isotropic as one explores deeper layers. This could partially be explained by the fact that the batch normalization \citep{pmlr-v37-ioffe15} used in some of these models mitigates \textit{a posteriori} the drift effect by removing the mean component of the representations. However, the ConvNeXt model also seems to use isotropic representations while not using batch normalization, which shows that this is not the only factor in the isotropic behavior of these models.

\subsection{To drift or not to drift?}
Related works \citep{bis-etal-2021-much, GaoHTQWL19} show that anisotropy in subword-level language models is caused by a drift of the hidden representations in a shared direction. In this section, we try to extend this observation to other modalities.

We study the correlation between the uniformly measured cosine-similarity, and the norm of the average hidden representation $||\bar{x}||_2$ for each layer. If anisotropy could be directly explained by the drift effect, we would expect a monotonic relation between $||\bar{x}||_2$ and the average cosine-similarity. To verify this, we apply a Spearman correlation test on these two metrics for every model from \autoref{sec:charbased} and \autoref{sec:other_mod}, along with some token-level language models, namely T5 \citep{2020t5}, BERT \citep{devlin-etal-2019-bert}, RoBERTa \citep{roberta}, and GPT-2 \citep{gpt2}.

\begin{figure}[h]
    \centering
    \includegraphics[width=0.9\linewidth]{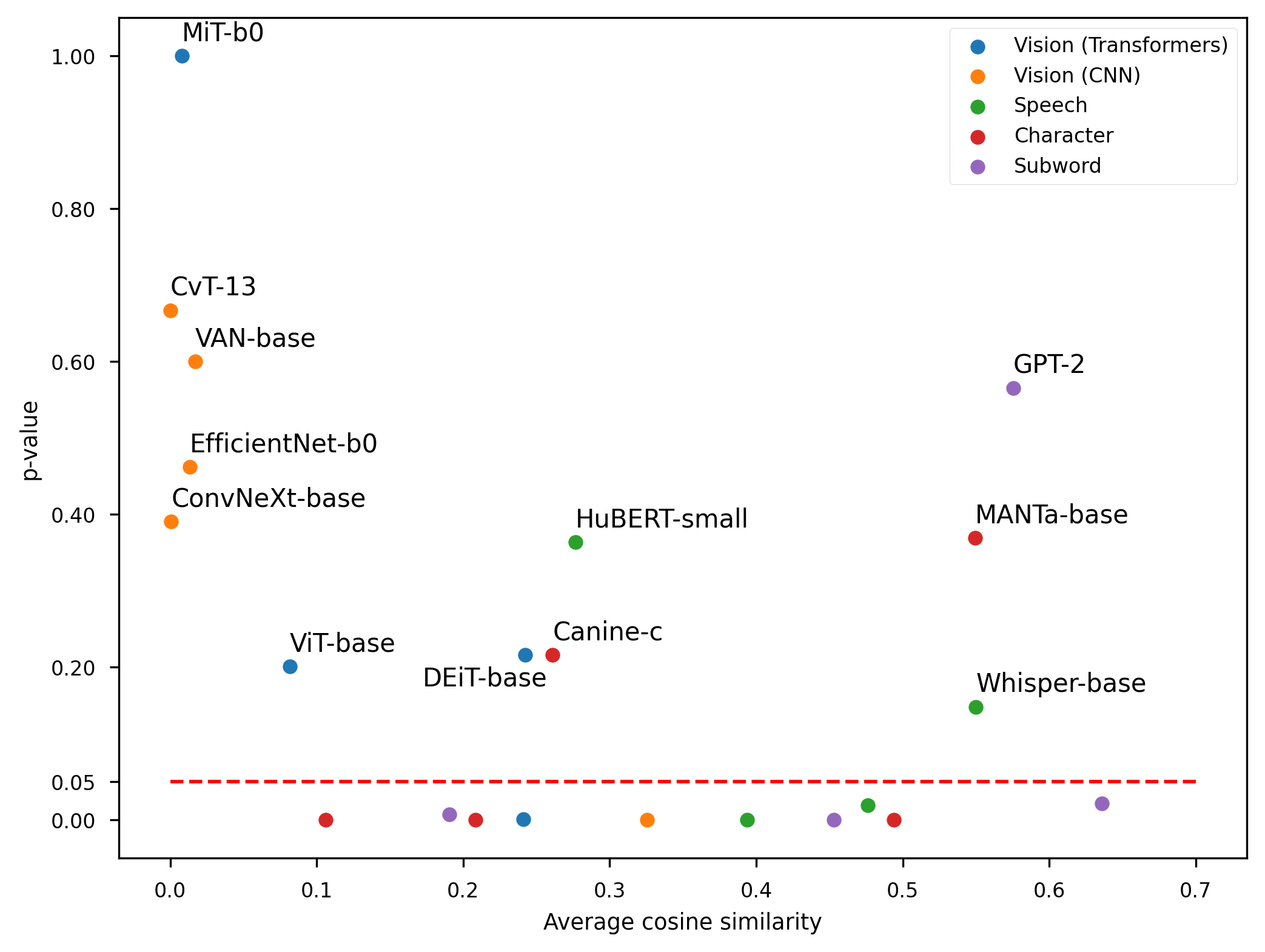}
    \caption{p-value of the Spearman correlation test between the norm of the average representation and the cosine-similarity averaged over all layers, across modalities. For models above the red dotted line, there is no significant ($p>0.05$) correlation between the drift effect and the anisotropy level.}
    \label{fig:pval_vs_cos}
\end{figure}

In \autoref{fig:pval_vs_cos}, we observe that we can correlate the anisotropy level and the magnitude of the drift component across layers for several models. The anisotropy of subword-based models can generally be correlated with the drift effect, except for GPT-2 for which the Spearman correlation metric may not be appropriate. We provide a similar analysis based on the Pearson correlation test and discuss the relevance of each statistic in \Cref{app:pearson}. 

Interestingly, we notice that the anisotropy affecting most CNN-based vision models is generally not correlated with the drift effect, contrary to Tranformers-based models in the same modality. Some speech models (HuBERT and Whisper-base) also display signs of anisotropy that cannot be correlated with the drift effect. \autoref{fig:pval_vs_cos} also shows a correlation for all character-based models but Canine-C and MANTa-base.

\section{Exploring the representation drift}
\label{sec:empirical}
In this section, we focus on some intrinsic properties of the Transformer block in a modality-agnostic fashion, i.e. with minimal assumptions on the data distribution, and without training. We analyze experimentally the behavior of the untrained Transformer block $T$ when a common bias term $b$ is added to untrained input representations $\mathbf{x}$. This allows us to mimic the common drift as mentioned in \citet{bis-etal-2021-much} and to identify some properties induced by this artificial drift on the output representations.

\subsection{Experimental setup}
We consider an embedding lookup table $E$ and a Transformer block $T$ with weights initialized as in BERT \citep{devlin-etal-2019-bert}. We then draw 16 input embedding sequences $\mathbf{x}$ of length 512 uniformly from $E$. To account for a drift component of norm $N\in\mathbb{R}$, we generate a vector $b_u \sim \mathcal{N}(0, I_d)$, which we normalize into $b = \frac{b_u}{||b_u||_2}\times N$. We finally compute $T(\mathbf{x}_i + b)$ for every sequence $x_i$, and study the resulting distributions.

Specifically, we study the average norm of the input representations $\mathbb{E}(||\mathbf{x}_i + b||_2)$ against the average norm of the output representations $\mathbb{E}(||T(\mathbf{x}_i + b)||_2)$ in \autoref{fig:norm_scratch_transformer}. We also retrieve the self-attention scores before the softmax operation, namely $\frac{QK^T}{\sqrt{d_k}}$, along with the corresponding $Q$ and $K$ matrices. We study some of their properties in \autoref{fig:attscore_trained_transformer} and \autoref{fig:kq}.

\subsection{Input vs. output analysis}
\begin{figure}[h]
    \centering
    \begin{subfigure}[b]{0.8\columnwidth}
         \includegraphics[width=\linewidth]{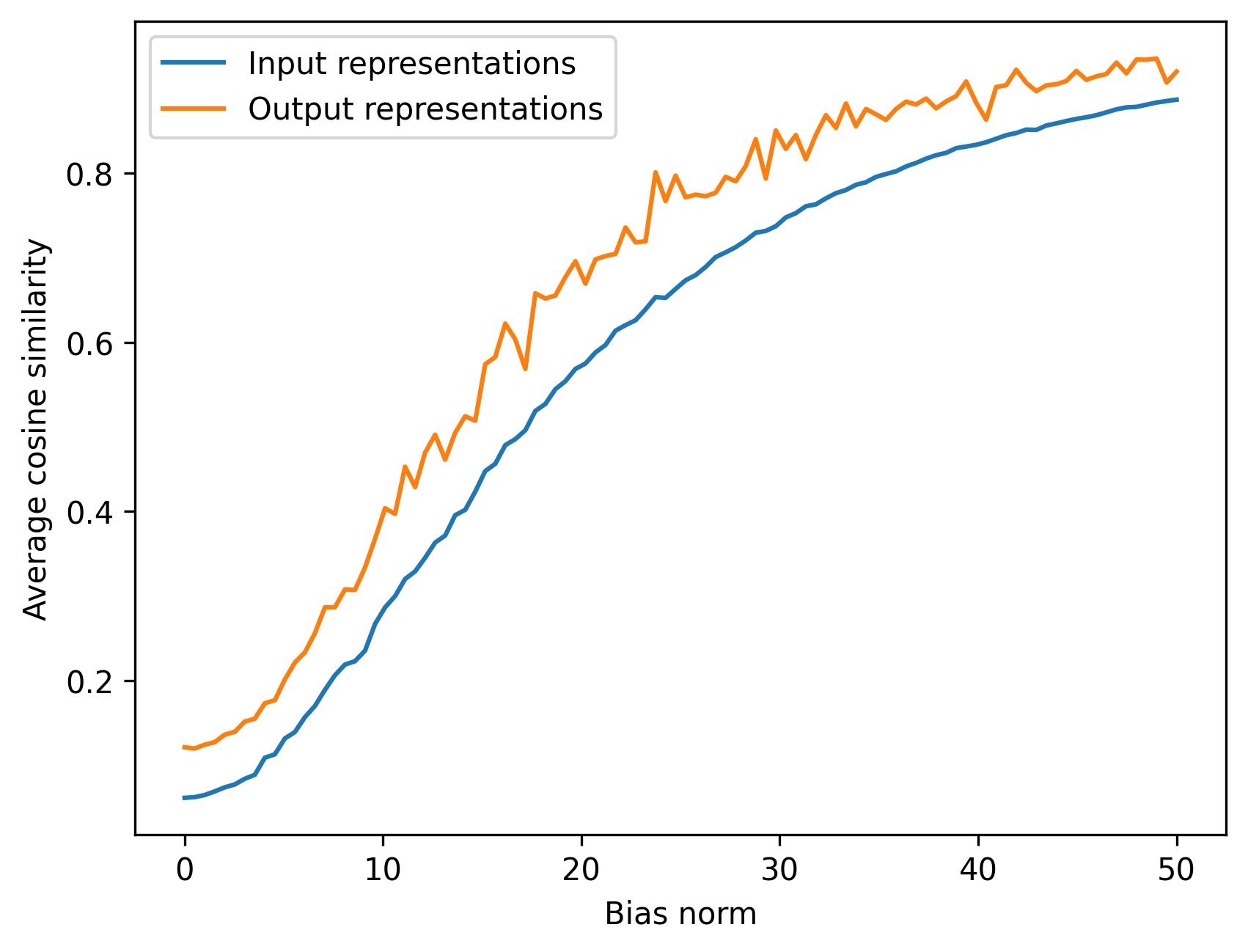}
         \subcaption{Cosine similarity}
         \label{fig:cos_scratch_transformer}
         
    \vspace{1.2em}
    \end{subfigure}
    \begin{subfigure}[b]{0.8\columnwidth}
         \includegraphics[width=\linewidth]{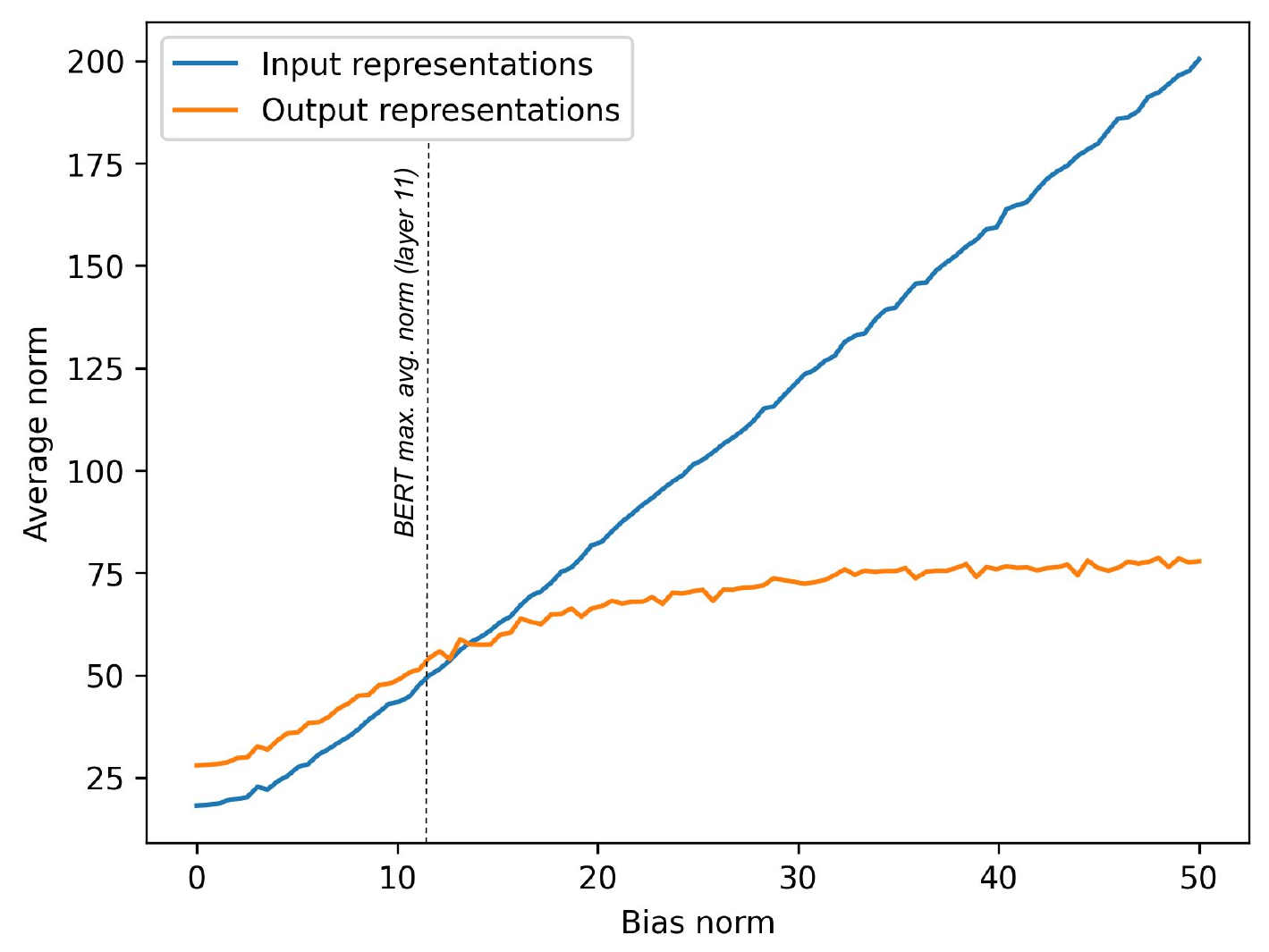}
         \subcaption{Norm}
         \label{fig:norm_scratch_transformer}
    \end{subfigure}
    \caption{Input/Output comparison of a Transformer block from BERT-base as the bias norms increases.}
    \label{fig:bias_vs_cosine_norm}
\end{figure}

In \autoref{fig:cos_scratch_transformer}, we observe that the output representations have an average cosine-similarity value that is slightly higher than the one of the input representations, no matter the level of input bias. We also notice that while the norm of the average output representation increases with the bias norm, it seems to meet the corresponding input measure for a given bias norm.

Interestingly, this shows that there is a \textit{fixed point} in terms of norm in the Transformers function with biased input. More formally, there seems to exist a bias norm $N^* \in \mathbb{R}_+$ such that: $$\mathbb{E}_{x, b_{N^*}}(||x_i + b_{N^*}||) = \mathbb{E}_{x, b_{N^*}}(||T(x_i + b_{N^*})||)$$

Moreover, this fixed point level $N^*$ is in the order of magnitude of the average hidden state norms of the layers of the trained BERT model. This hints that the model's representations stabilize when their norm is close to this fixed point. We leave a more thorough analysis of this hypothesis for future work.

\subsection{Exploring the Transformer block}

To understand the effect of the drift effect on the inner workings of the Transformer layer, we take a closer look at the self-attention operation as the average input representation drifts away.

\begin{figure}[h]
    \centering
    \includegraphics[width=0.9\linewidth]{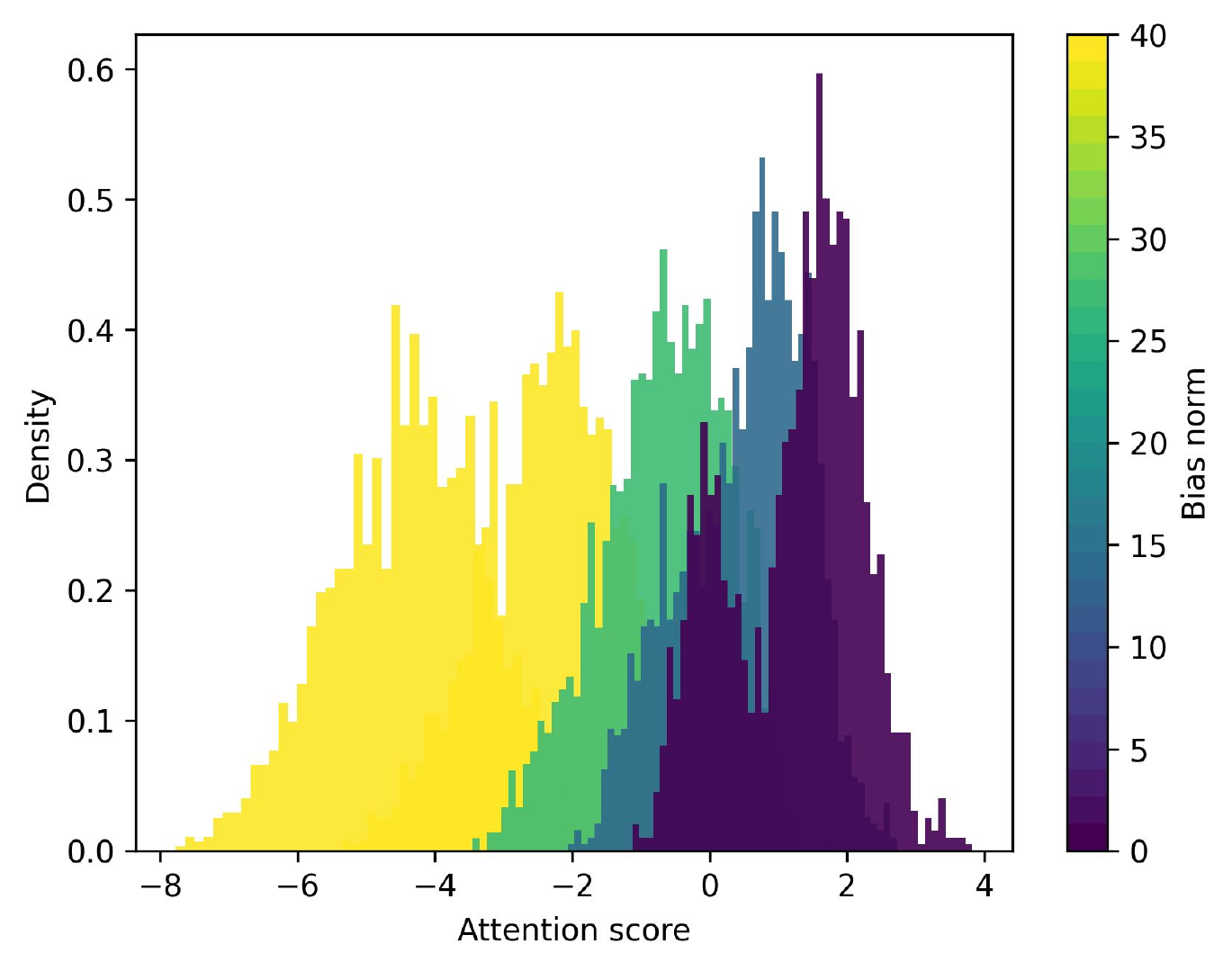}
    \caption{Histograms of the pre-softmax attention scores as the input bias norm increases. Other initializations of the layer and of the bias direction $b_u$ led to a general \textit{increase} of the attention scores instead.}
    \label{fig:attscore_trained_transformer}
\end{figure}

\autoref{fig:attscore_trained_transformer} shows that the attention scores tend to move away from zero as the input bias norm increases. Indeed, as the norm of the average $\bar{x}$ of the input embeddings increases, we can expect the query and key vectors $Q$ and $K$ to also display signs of anisotropy. Actually, for each self-attention head, and for all position $i \in [1, L]$, we have:
\begin{equation}
    \begin{cases}
      \mathbb{E}_x(Q_i) = W_Q\bar{x} + b_Q\\
      \mathbb{E}_x(K_i) = W_K\bar{x} + b_K
    \end{cases}
\end{equation}

\begin{figure}[h]
    \centering
    \begin{subfigure}[b]{0.48\columnwidth}
         \includegraphics[width=\linewidth]{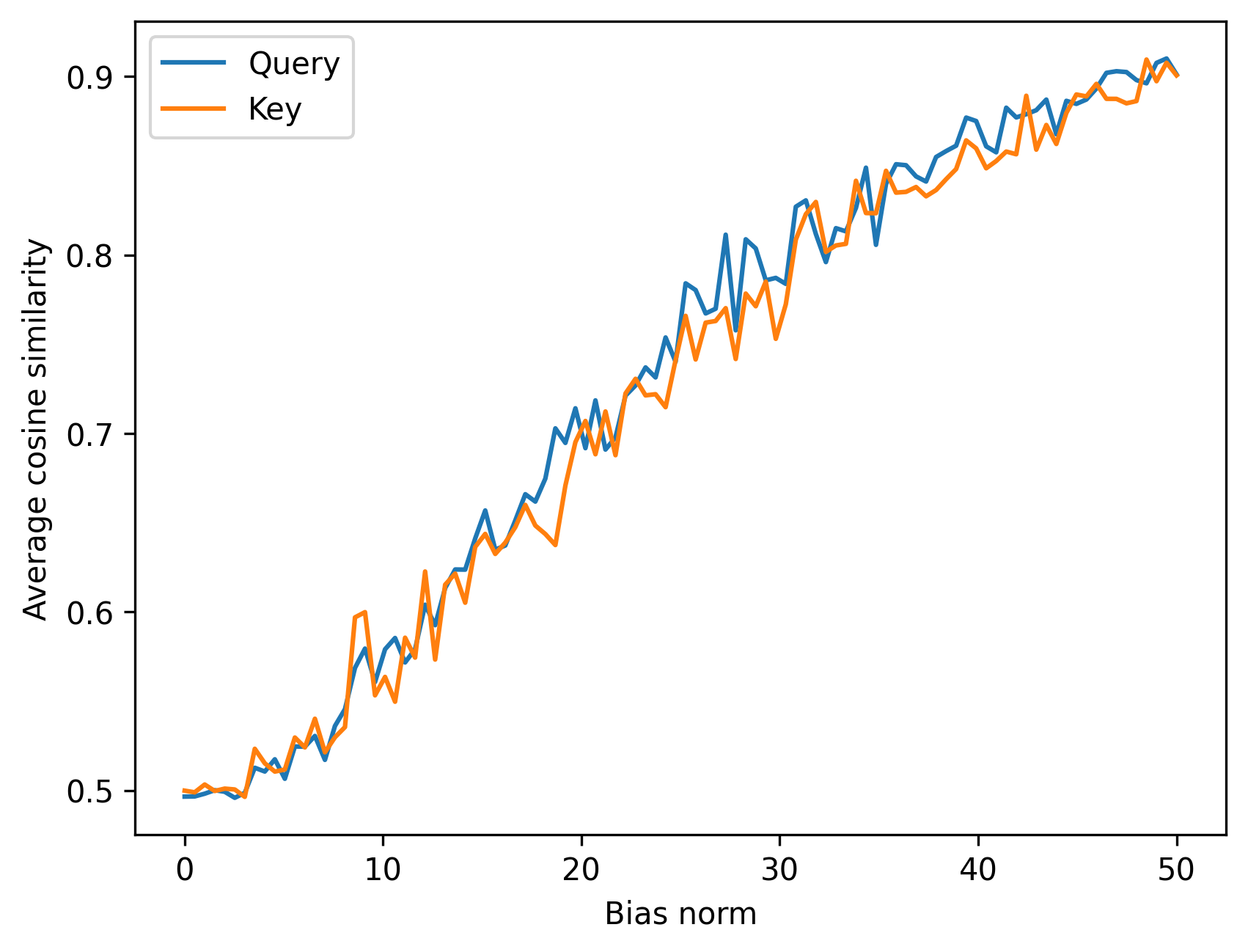}
         \caption{Cosine sim.}
         \label{fig:cos_qk_trained_transformer}
    \end{subfigure}
    \begin{subfigure}[b]{0.48\columnwidth}
         \includegraphics[width=\linewidth]{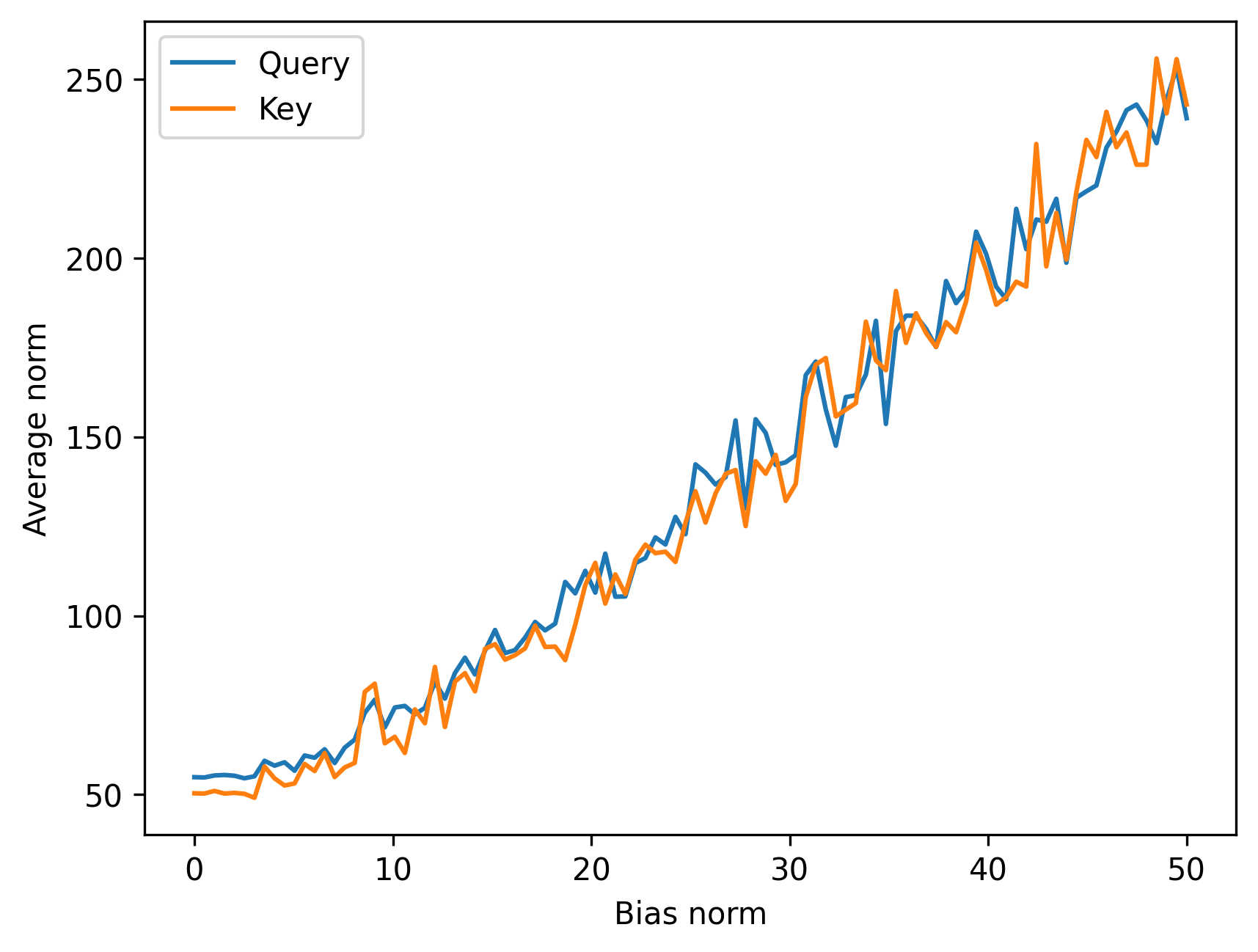}
         \caption{Norm}
         \label{fig:norm_qk_trained_transformer}
    \end{subfigure}
    \caption{Analysis of the self-attention query and key distributions}
    \label{fig:kq}
\end{figure}

We can observe in \autoref{fig:kq} that query and key representations indeed increase in norm with the input bias norm. We also notice that the corresponding distributions are anisotropic even when no bias is added, which may be a consequence of BERT's initialization parameters.

\subsection{Impact of the drift}

After exploring the consequences of the drift of input representations on the query-key product in self-attention, we identify in this section the implications of this drift at a more explainable level, by observing the resulting post-softmax distributions.

\begin{figure}[h]
    \centering
    \includegraphics[width=0.8\linewidth]{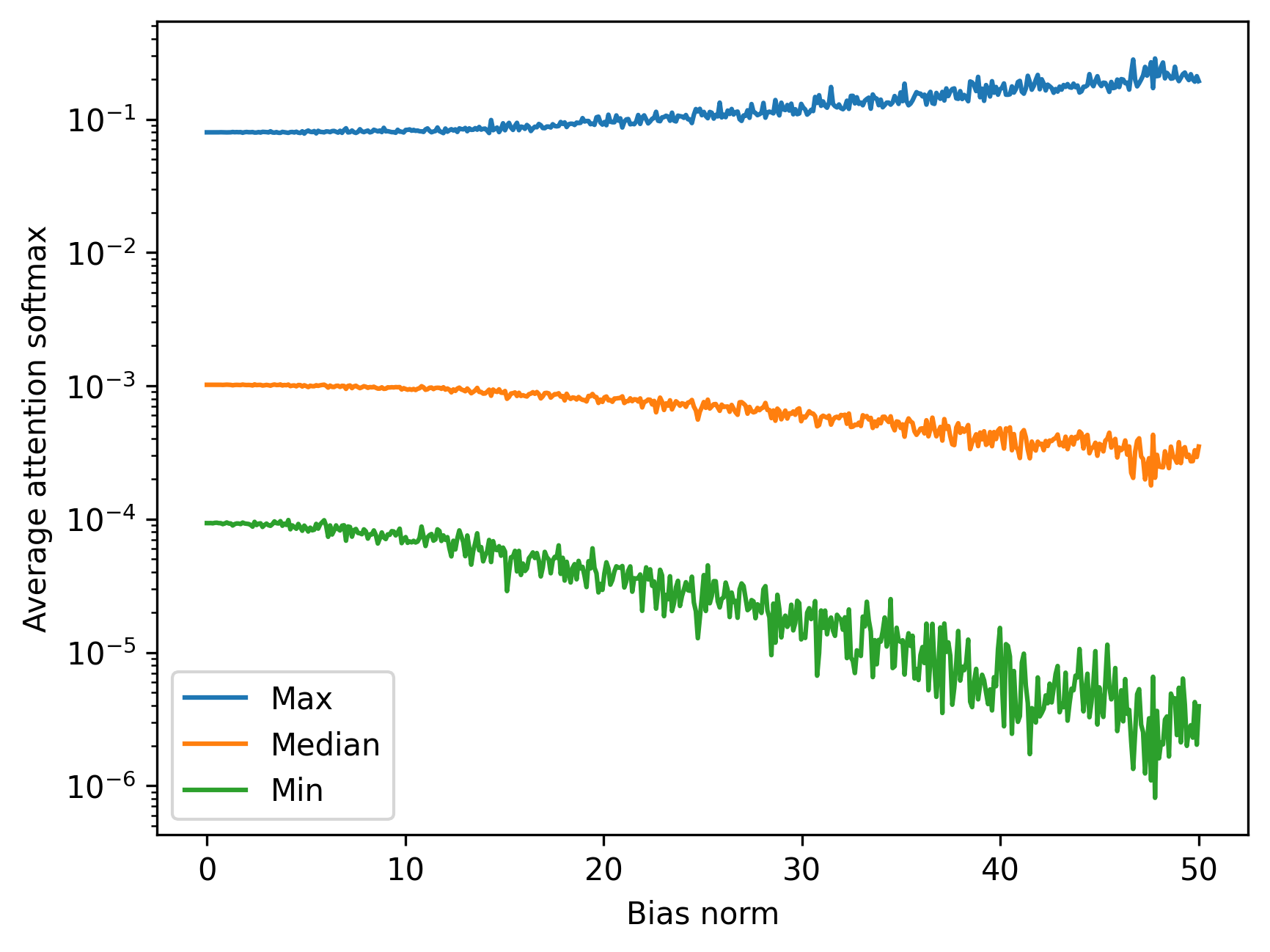}
    \caption{Evolution of the self-attention softmax values as the input bias norm increases.}
    \label{fig:softmax_trained_transformer}
\end{figure}

In \autoref{fig:softmax_trained_transformer}, we retrieve softmax values in the self-attention block and for each position, we extract the maximum, the median and the minimum. We then average these values over the whole batch, and repeat for various input bias norm levels. We notice that as the input bias norm increases, the self-attention softmax distributions tend to become less entropic, evolving towards higher maximal probabilities and lower minimal probabilities. In the following analysis, we'll use the term \textit{sharpness} to discuss entropy levels of the self-attention distributions.

\begin{figure}[h]
    \centering
    \begin{subfigure}[b]{0.48\columnwidth}
         \includegraphics[width=\linewidth]{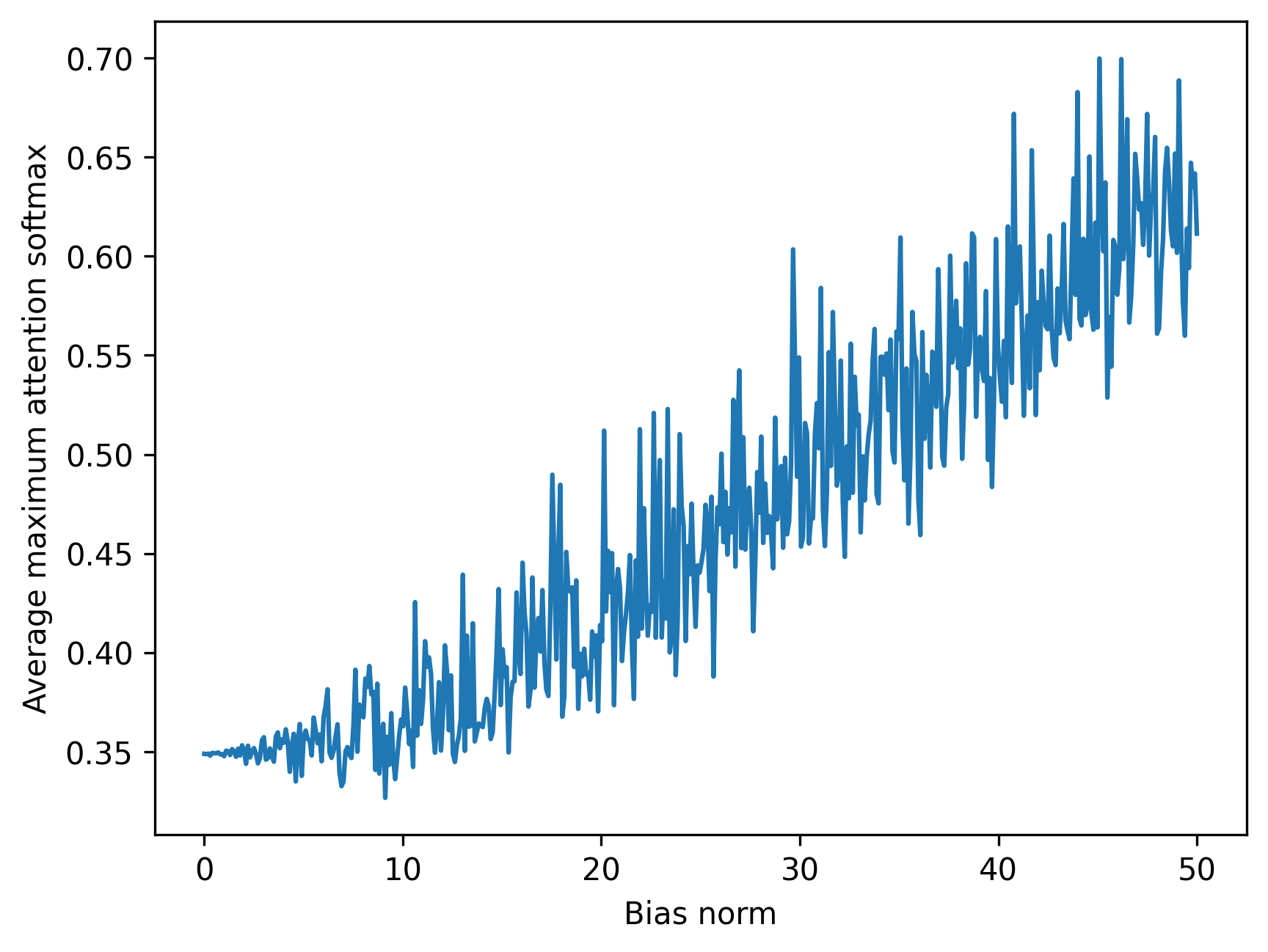}
         \caption{Maximum}
         \label{fig:max_softmax}
    \end{subfigure}
    \begin{subfigure}[b]{0.48\columnwidth}
         \includegraphics[width=\linewidth]{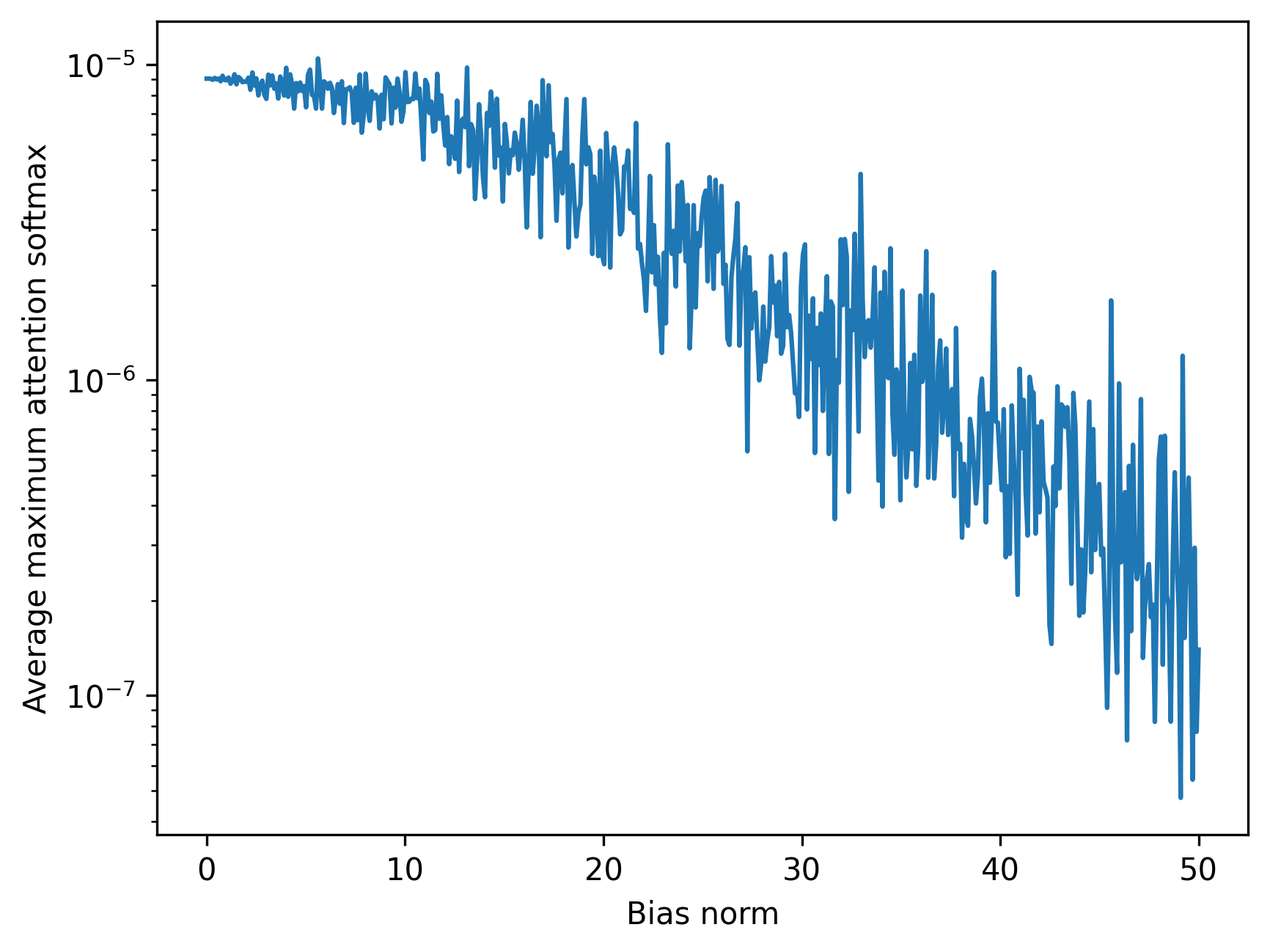}
         \caption{Minimum}
         \label{fig:min_softmax}
    \end{subfigure}
    \caption{Comparison of the extreme values of each sequence averaged over the batch as the bias norm increases.}
    \label{fig:min_vs_max}
\end{figure}

This sharpening effect of the attention distributions becomes even clearer if we consider the maximum and minimum values over the whole sequences, as in \autoref{fig:min_vs_max}.

However, at low anisotropy levels, i.e. when the bias norm is low, we see that the effect is not very important. \autoref{fig:softmax_trained_transformer} and \autoref{fig:min_vs_max} only hint at the fact that the drift of embeddings may help the self-attention to be sharper. Another explanation could be that training favors sharp self-attention patterns, as has been pointed out in previous works \citep{clark-etal-2019-bert}, which in turn induces a drift in the models' representations. In order to account for that, we need to study the evolution of latent spaces at the self-attention level along training.

\section{Queries and keys: training dynamics}
\label{sec:qk}
\begin{figure*}[h]
    \centering
    \begin{subfigure}[b]{0.24\linewidth}
         \includegraphics[width=\linewidth]{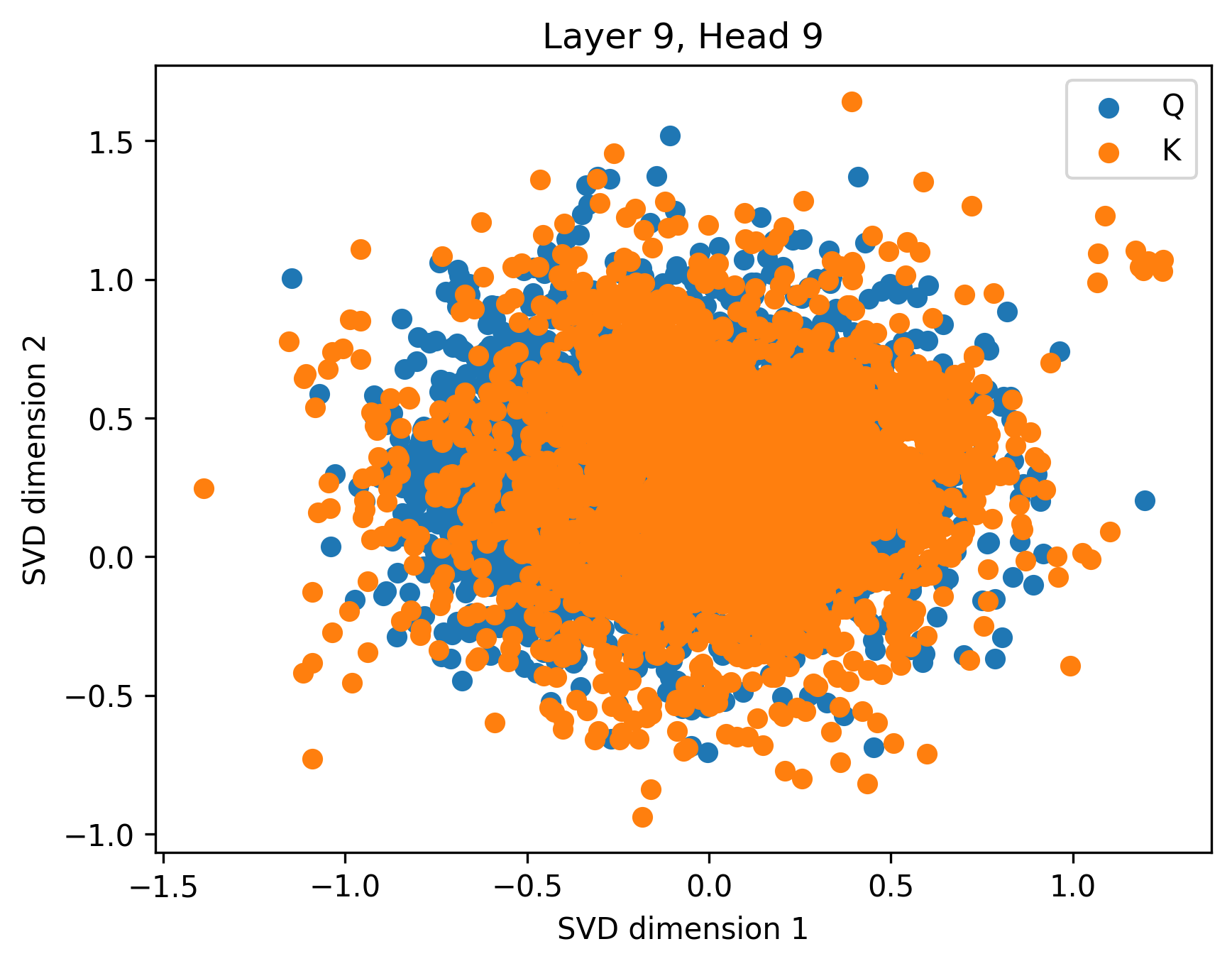}
         \caption{Step 0}
         \label{fig:dist_qk_s0}
    \end{subfigure}
    \begin{subfigure}[b]{0.24\linewidth}
         \includegraphics[width=\linewidth]{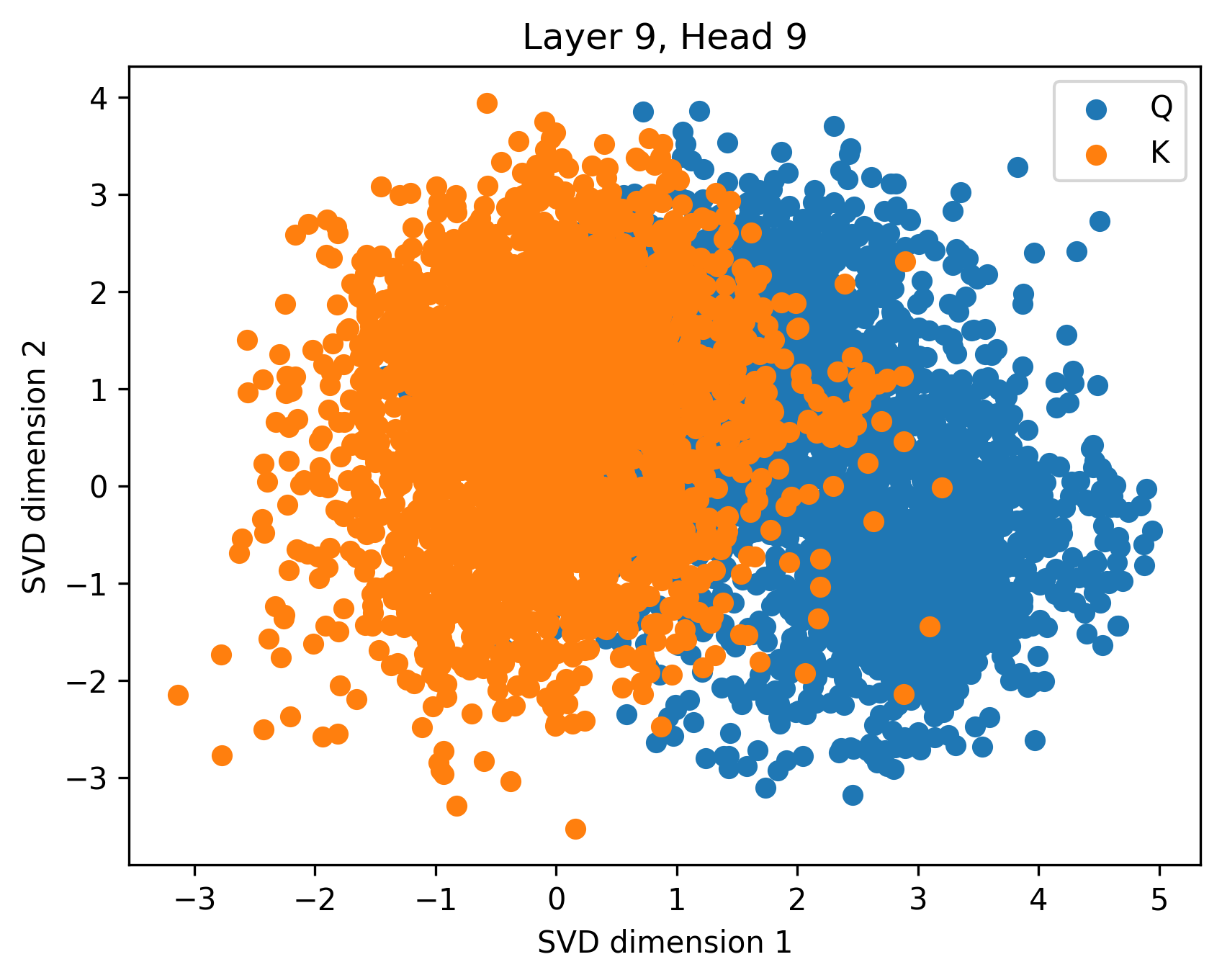}
         \caption{Step 40k}
         \label{fig:dist_qk_s40}
    \end{subfigure}
    \begin{subfigure}[b]{0.24\linewidth}
         \includegraphics[width=\linewidth]{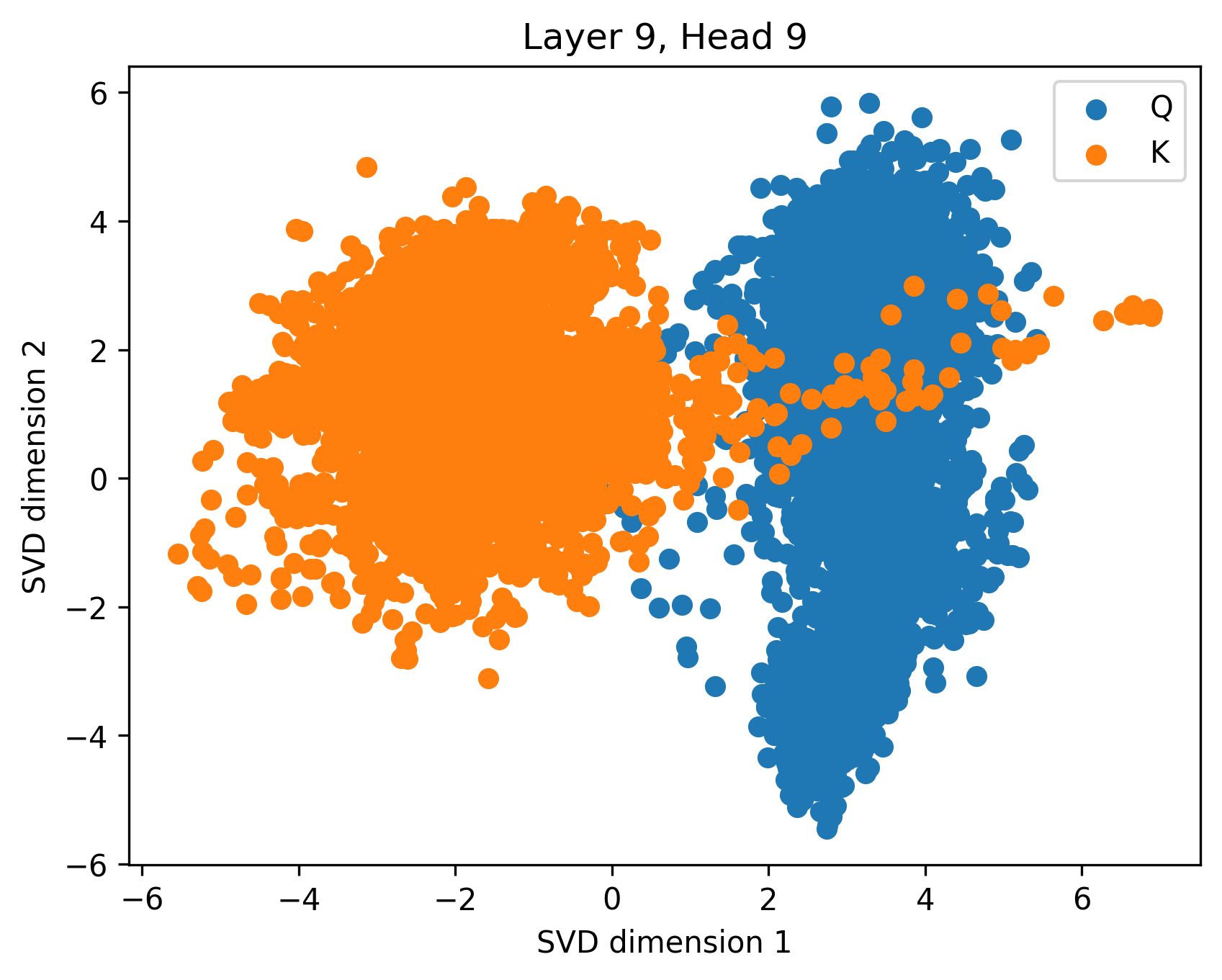}
         \caption{Step 200k}
         \label{fig:dist_qk_s200}
    \end{subfigure}
    \begin{subfigure}[b]{0.24\linewidth}
         \includegraphics[width=\linewidth]{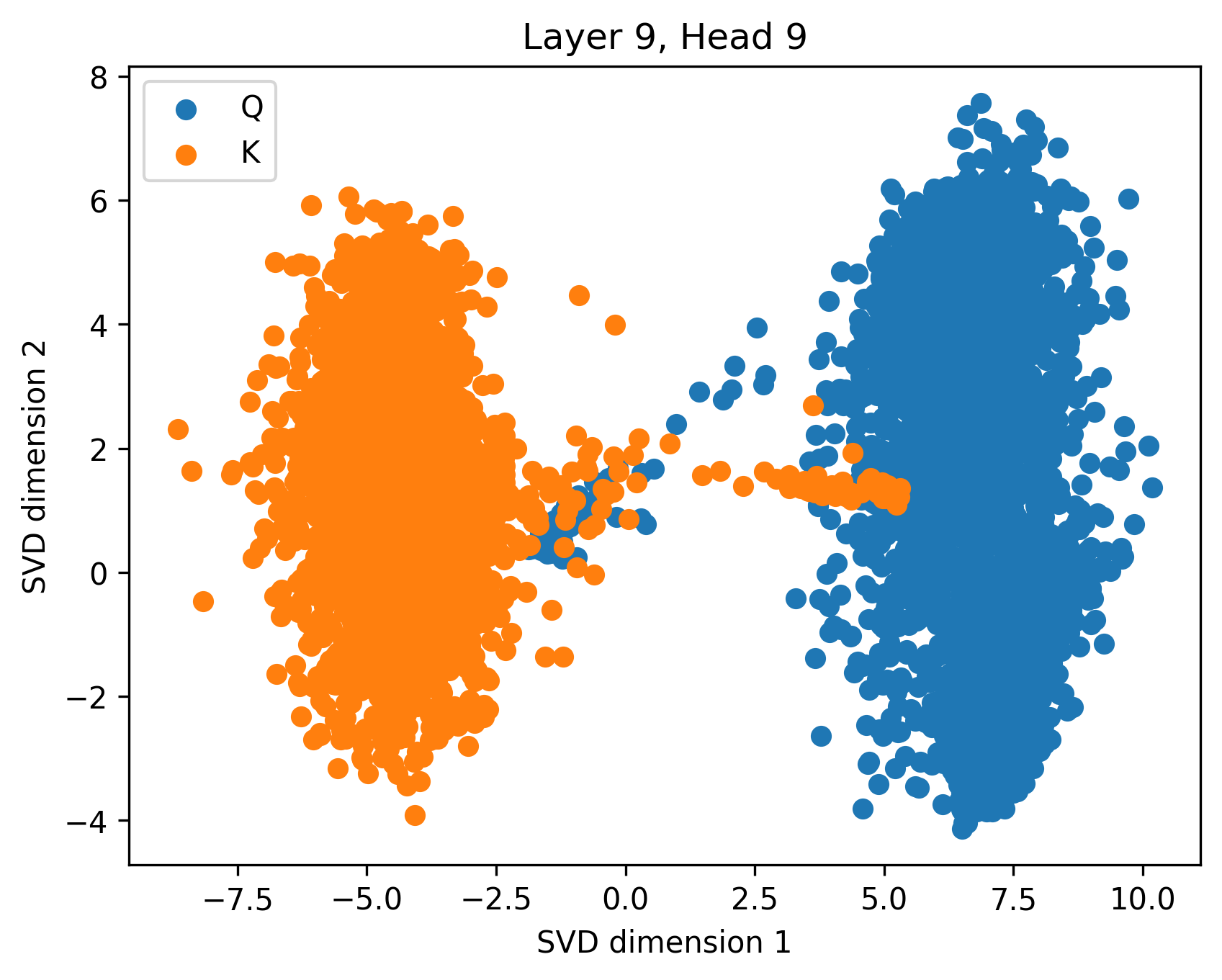}
         \caption{Step 2M (final)}
         \label{fig:dist_qk_s2M}
    \end{subfigure}
    \caption{Evolution of $Q_s$ and $K_s$ distributions along training. Vectors are projected using a common SVD.}
    \label{fig:proj_qk_heads}
\end{figure*}

We have established that manually pushing for drift-based anisotropy on \textit{untrained} Transformers models leads to sharper (i.e. low-entropy) self-attention patterns. In this section, we show that this evolution of self-attention values actually takes place during training, and we explore the mechanism behind their appearance. As pointed out in \autoref{sec:empirical}, the self-attention scores result from the $QK^T$ operation, which computes scalar products between query and key representations corresponding to each pair of positions. Thus, in this section, we study the evolution of these query and key representations \textit{along training}, and explore the mechanism behind the increase of the scalar products leading to self-attention scores.

We use the MultiBERT checkpoints \citep{sellam2021multiberts} with seed 0 to retrieve $Q$ and $K$ distributions at different pretraining steps, and we use 128 samples from Wikitext-103 as input data. Along this section, $Q_s$ and $K_s$ refer to query and key representations extracted at a specific layer and head at a given step $s$, and $\hat{Q_s}$ and $\hat{K_s}$ are the average representations, taken over all tokens in the sampled batch. By studying $\bar{Q_s}$ and $\bar{K_s}$, we aim at exploring the common (or context-agnostic) drifts of keys and queries distributions.

\begin{figure}[h]
    \centering
    \begin{subfigure}[b]{0.48\columnwidth}
         \includegraphics[width=\linewidth]{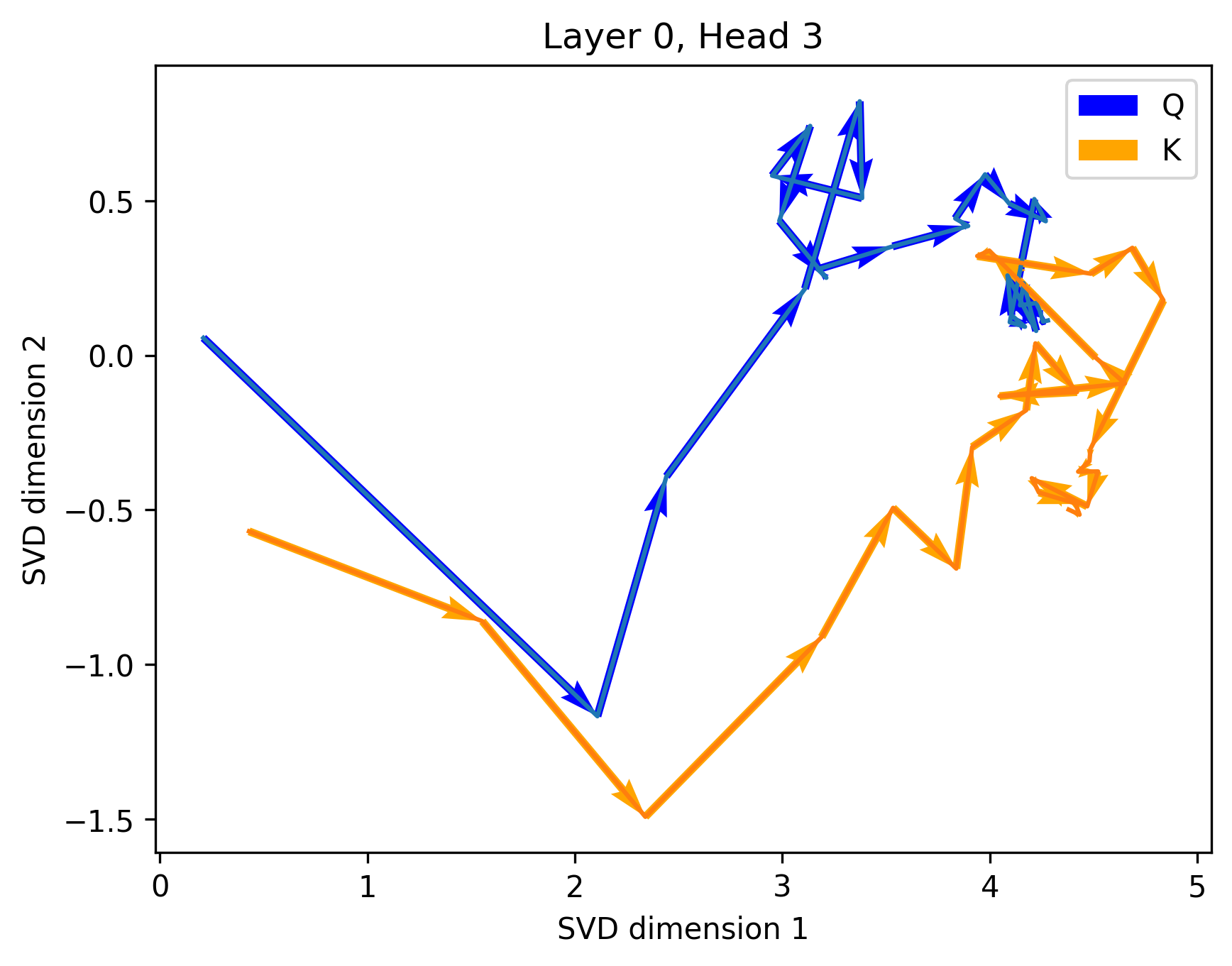}
         \caption{Similar}
         \label{fig:QK_simdir}
    \end{subfigure}
    \begin{subfigure}[b]{0.48\columnwidth}
         \includegraphics[width=\linewidth]{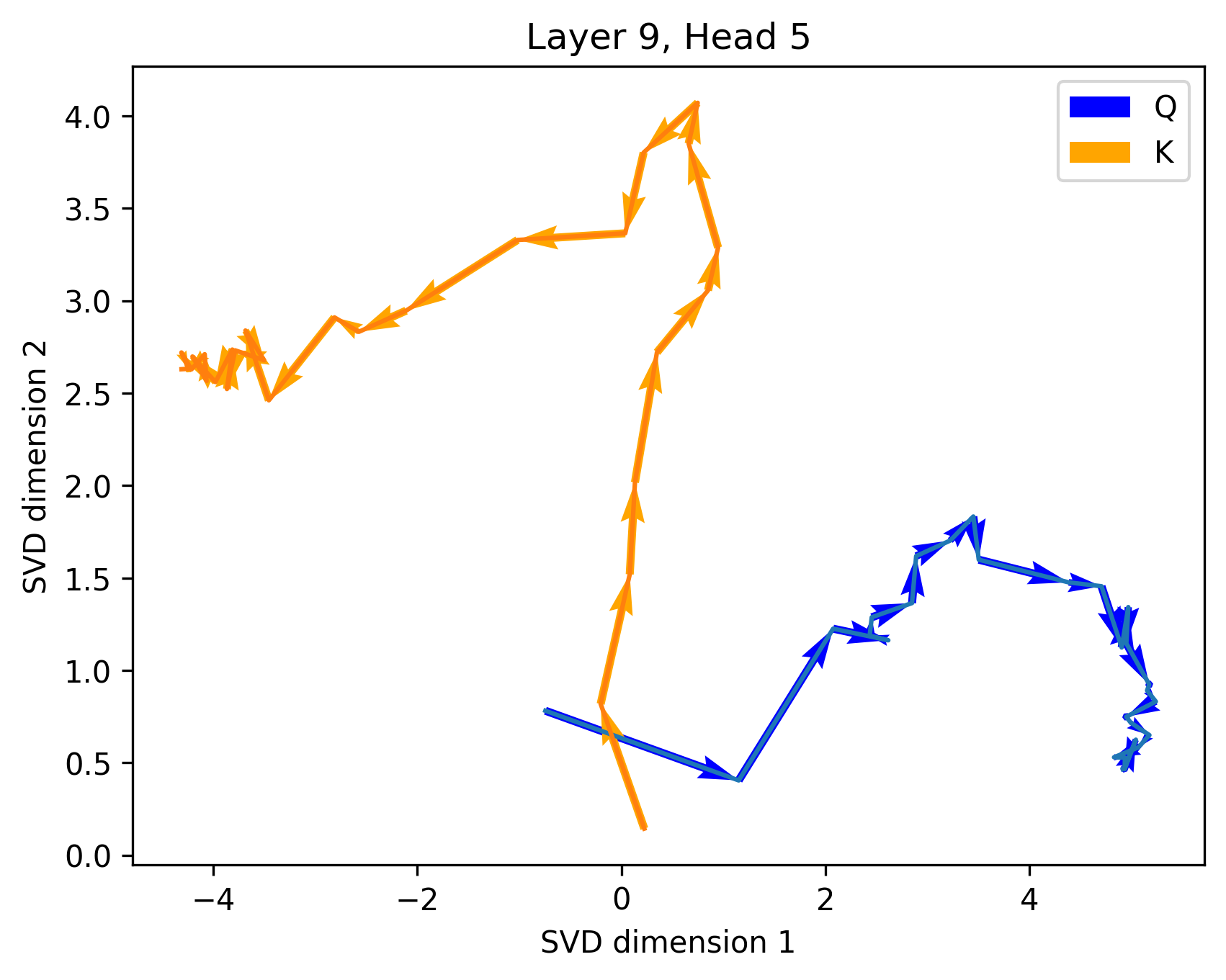}
         \caption{Opposite}
         \label{fig:QK_diffdir}
    \end{subfigure}
    \caption{Evolution of $\bar{Q_s}$ and $\bar{K_s}$ along training for two different heads in the network, projected via common SVD. Each arrow represents a checkpoint in the MultiBERT suite. We display typical examples of dynamics in same/opposite direction.}
    \label{fig:QK_dir}
\end{figure}

In \autoref{fig:proj_qk_heads} and \autoref{fig:QK_dir}, we compute a SVD of the union of $Q_s$ and $K_s$ for all steps $s$, so that the projection makes sense for both distributions across steps for visualization purposes \footnote{We actually uniformly sample 20\% of the whole set of representations to compute the SVD under reasonable memory constraints.}. As shown in our selected examples, we observe that the dynamics of $\bar{Q_s}$ and $\bar{K_s}$ tend to align along training, making the average of the distributions drift in either similar or opposite directions. The first dimension of the SVD seems to describe this common drift. Note that in $\mathbb{R}^{d_h}$ ($d_h = 64$ being the head dimension), such an alignment is very unlikely to happen randomly. Interestingly, \autoref{fig:QK_simdir} shows that the common direction dynamics appear in the first few steps, while the opposite direction dynamics of  \autoref{fig:QK_diffdir} only starts after 8\% of the total training steps.

\begin{figure*}[h]
    \centering
    \begin{subfigure}[b]{0.24\linewidth}
         \includegraphics[width=\linewidth]{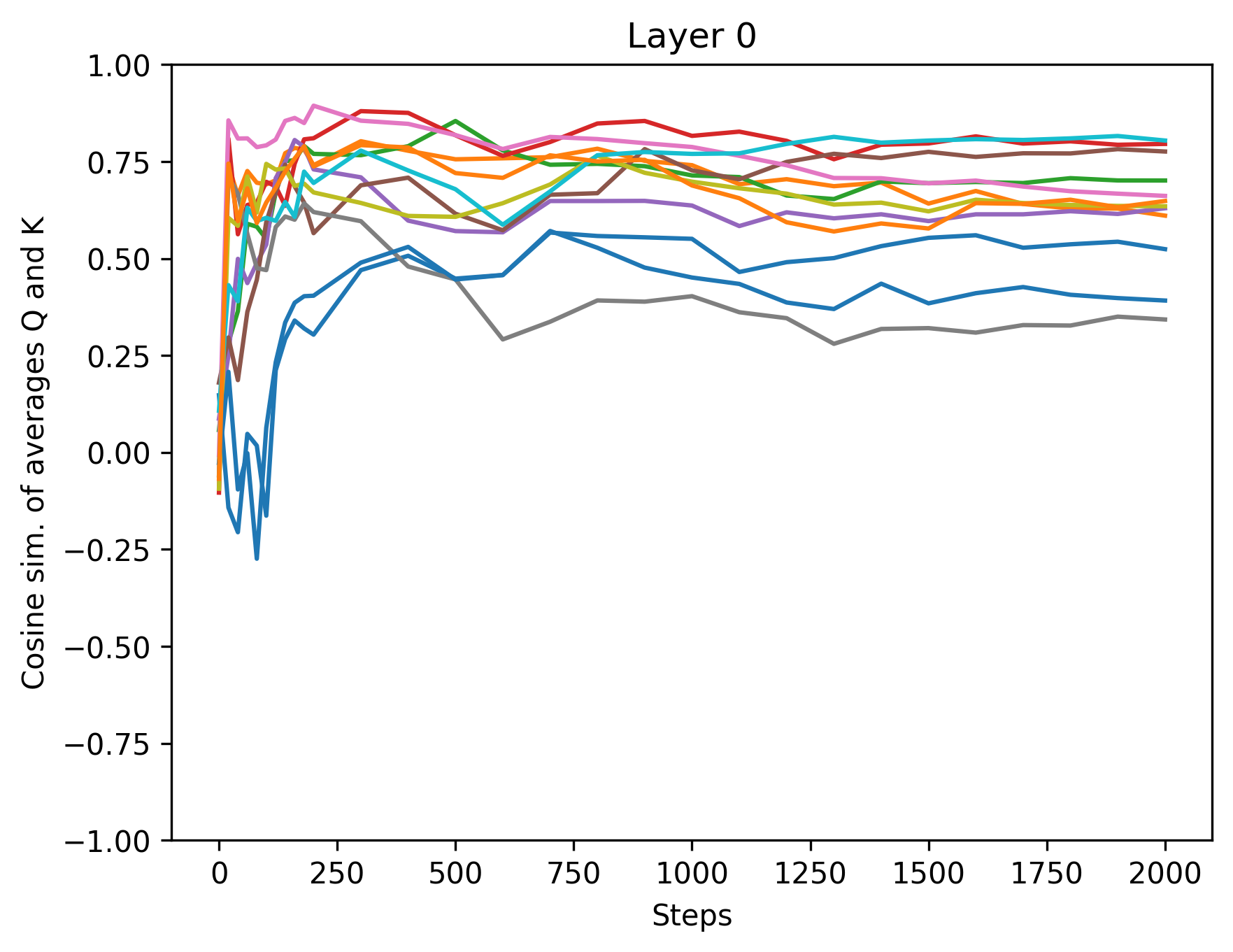}
         \caption{Layer 0}
         \label{fig:cosine_qk_l0}
    \end{subfigure}
    \begin{subfigure}[b]{0.24\linewidth}
         \includegraphics[width=\linewidth]{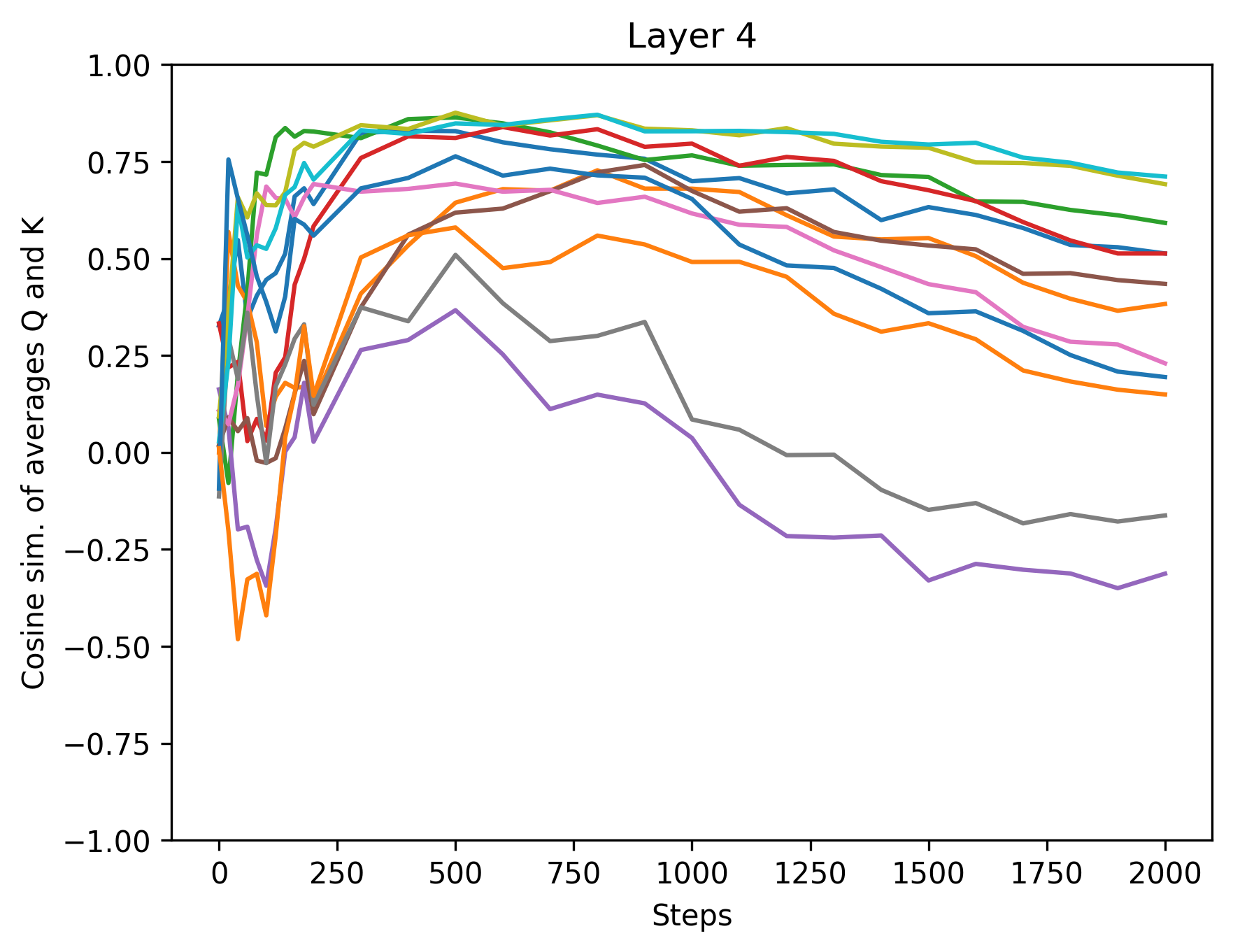}
         \caption{Layer 4}
         \label{fig:cosine_qk_l4}
    \end{subfigure}
    \begin{subfigure}[b]{0.24\linewidth}
         \includegraphics[width=\linewidth]{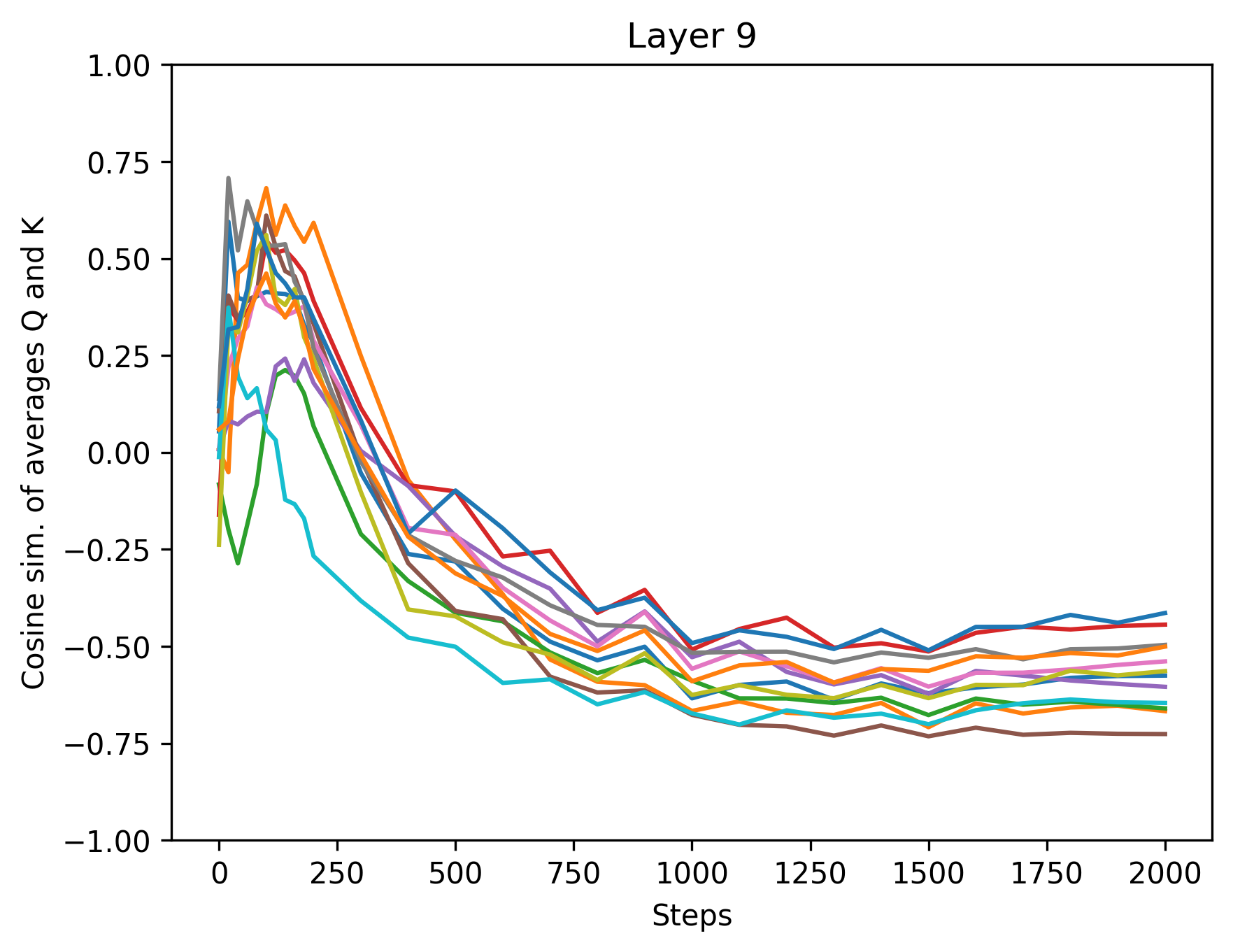}
         \caption{Layer 9}
         \label{fig:cosine_qk_l9}
    \end{subfigure}
    \begin{subfigure}[b]{0.24\linewidth}
         \includegraphics[width=\linewidth]{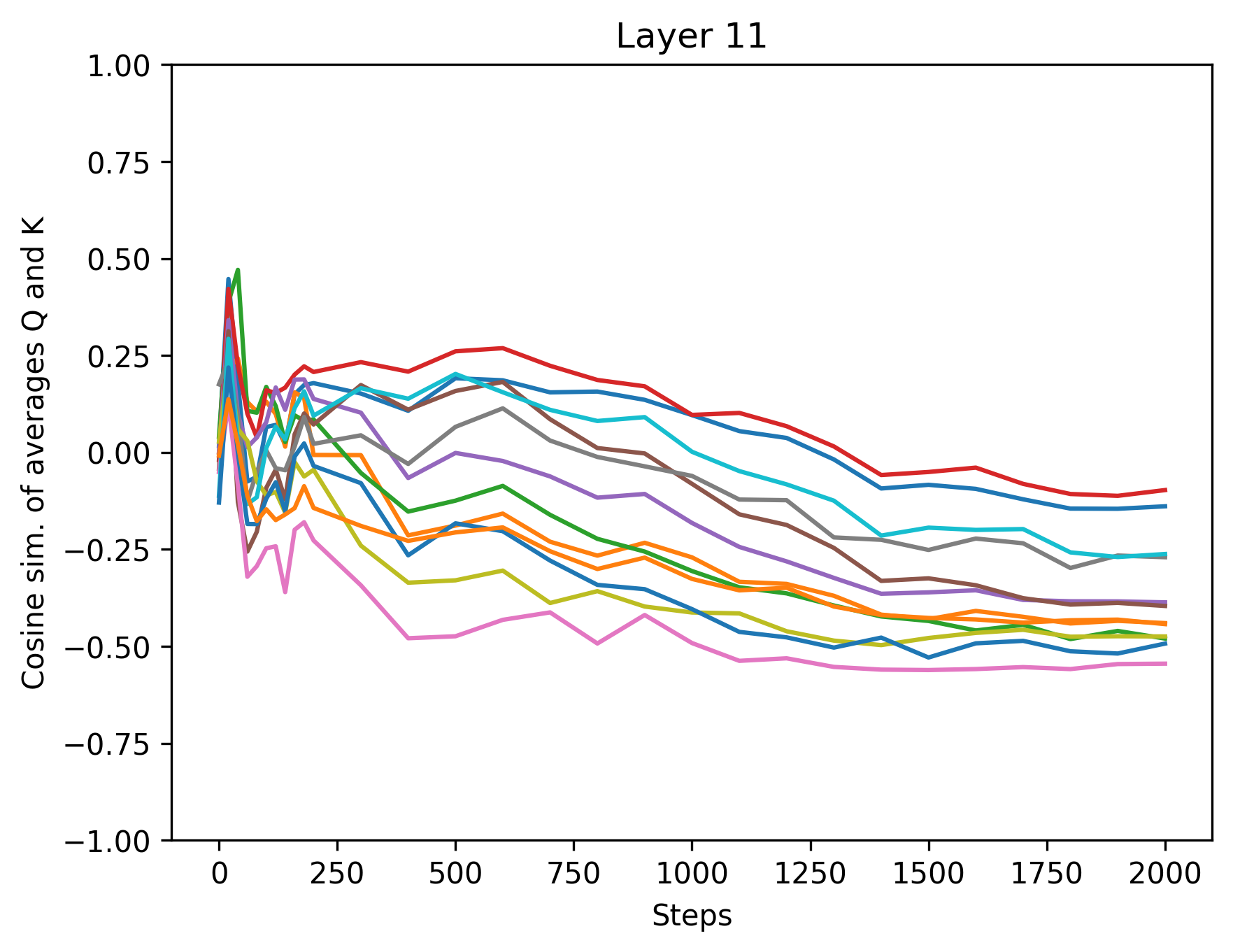}
         \caption{Layer 11}
         \label{fig:cosine_qk_l11}
    \end{subfigure}
    \caption{Evolution of cosine-similarity between $\bar{Q_s}$ and $\bar{K_s}$ along training. Each color represents one self-attention head. Steps are counted in thousands. We generally observe that almost all heads see $\bar{Q_s}$ and $\bar{K_s}$ align in common or opposite directions along training. In other words, the average components of keys and queries representations tend to align in self-attention heads, which maximizes the magnitude of the scalar product between two average representations. We run a similar experiment on all MultiBERT seeds in \autoref{fig:seeds_qk}, and obtain comparable results.}
    \label{fig:cosine_qk_heads}
\end{figure*}

To consolidate our observations, we compute the evolution of the cosine-similarity between $\bar{Q_s}$ and $\bar{K_s}$ along training in \autoref{fig:cosine_qk_heads}. We also display some projected $Q_s$ and $K_s$ distributions for several $s$ steps in \autoref{fig:proj_qk_heads}.

\autoref{fig:cosine_qk_heads} shows that the first layers display a common direction dynamic, as the cosine-similarity tends to increase, thus showing that \textbf{the key and query distributions drift along a similar direction} in average. The last layers seem to adopt an opposite direction dynamic, as the cosine-similarity between their mean key and query representations gets negative along training.

\begin{figure}[h]
    \centering
    \begin{subfigure}[b]{0.48\columnwidth}
         \includegraphics[width=\linewidth]{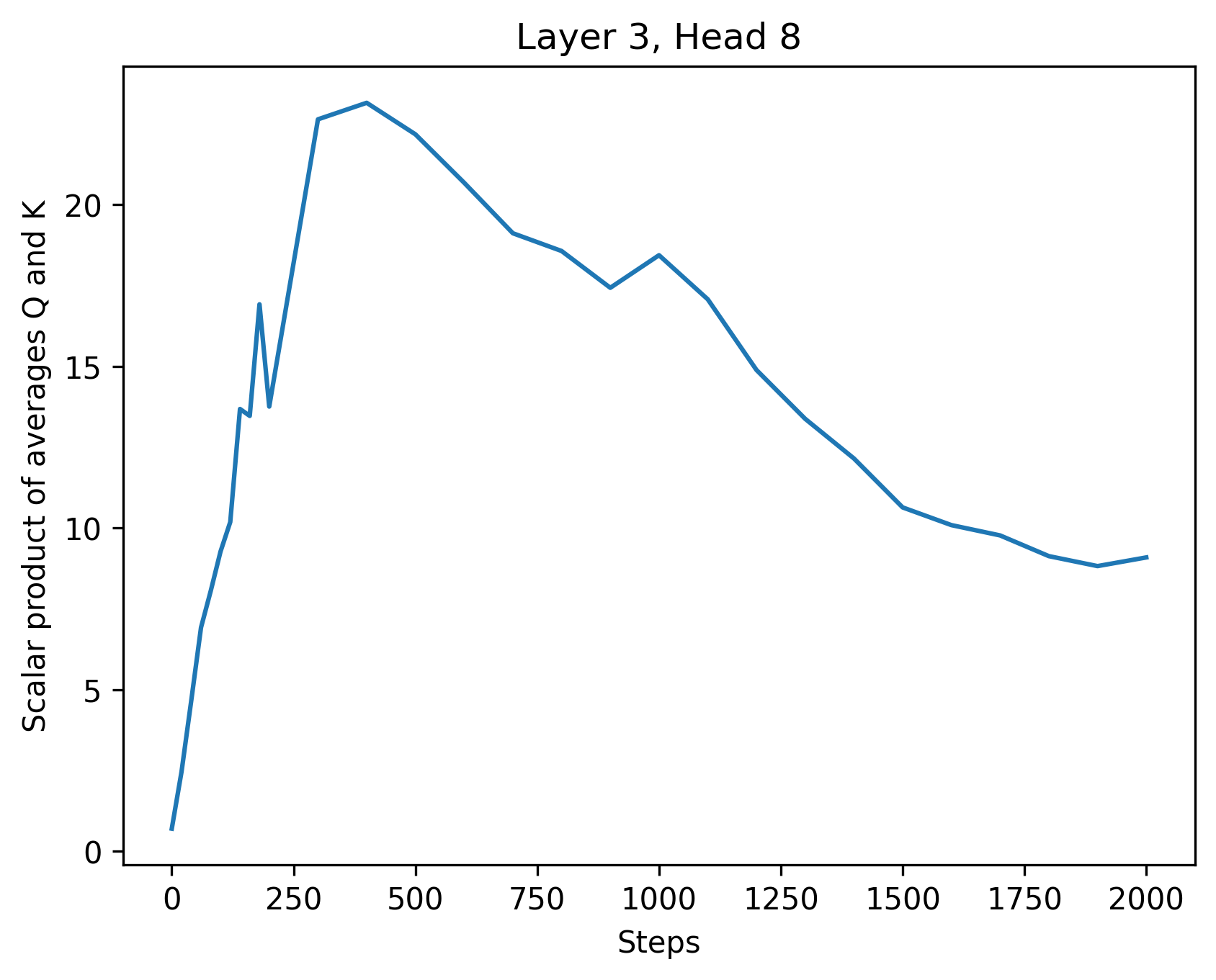}
         \caption{Similar}
         \label{fig:scalar_sim}
    \end{subfigure}
    \begin{subfigure}[b]{0.48\columnwidth}
         \includegraphics[width=\linewidth]{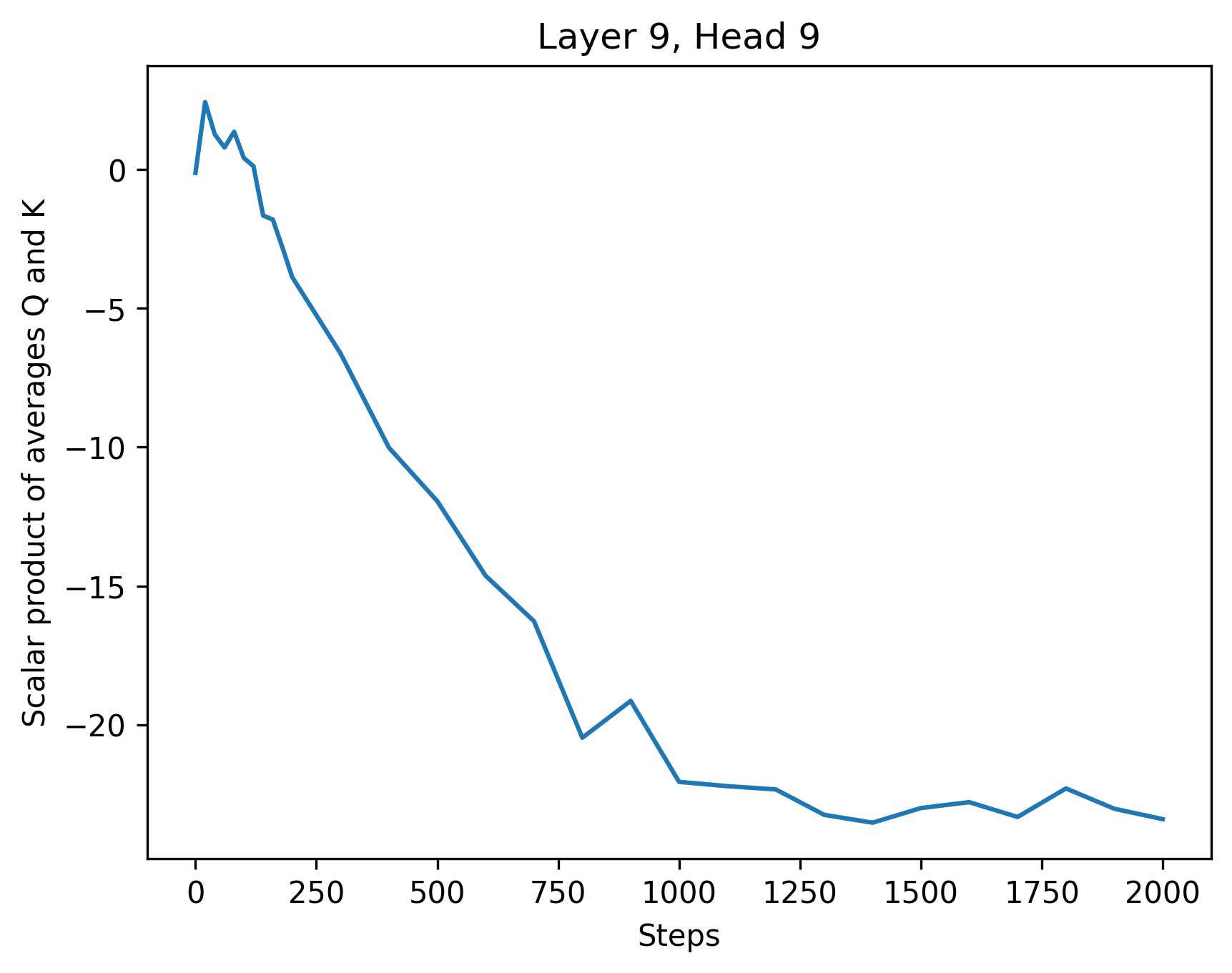}
         \caption{Opposite}
         \label{fig:scalar_opp}
    \end{subfigure}
    \caption{Evolution of the scalar product between $\bar{Q_s}$ and $\bar{K_s}$ along training. Steps are in thousands.}
    \label{fig:scalar_QK}
\end{figure}

As shown in \autoref{fig:scalar_QK}, this drift induces an increase in the magnitude of scalar products obtained in the self-attention $QK^T$ operation, thus facilitating the emergence of sharp patterns where attention focuses on specific tokens.

\begin{figure}[h]
    \centering
    \includegraphics[width=0.8\linewidth]{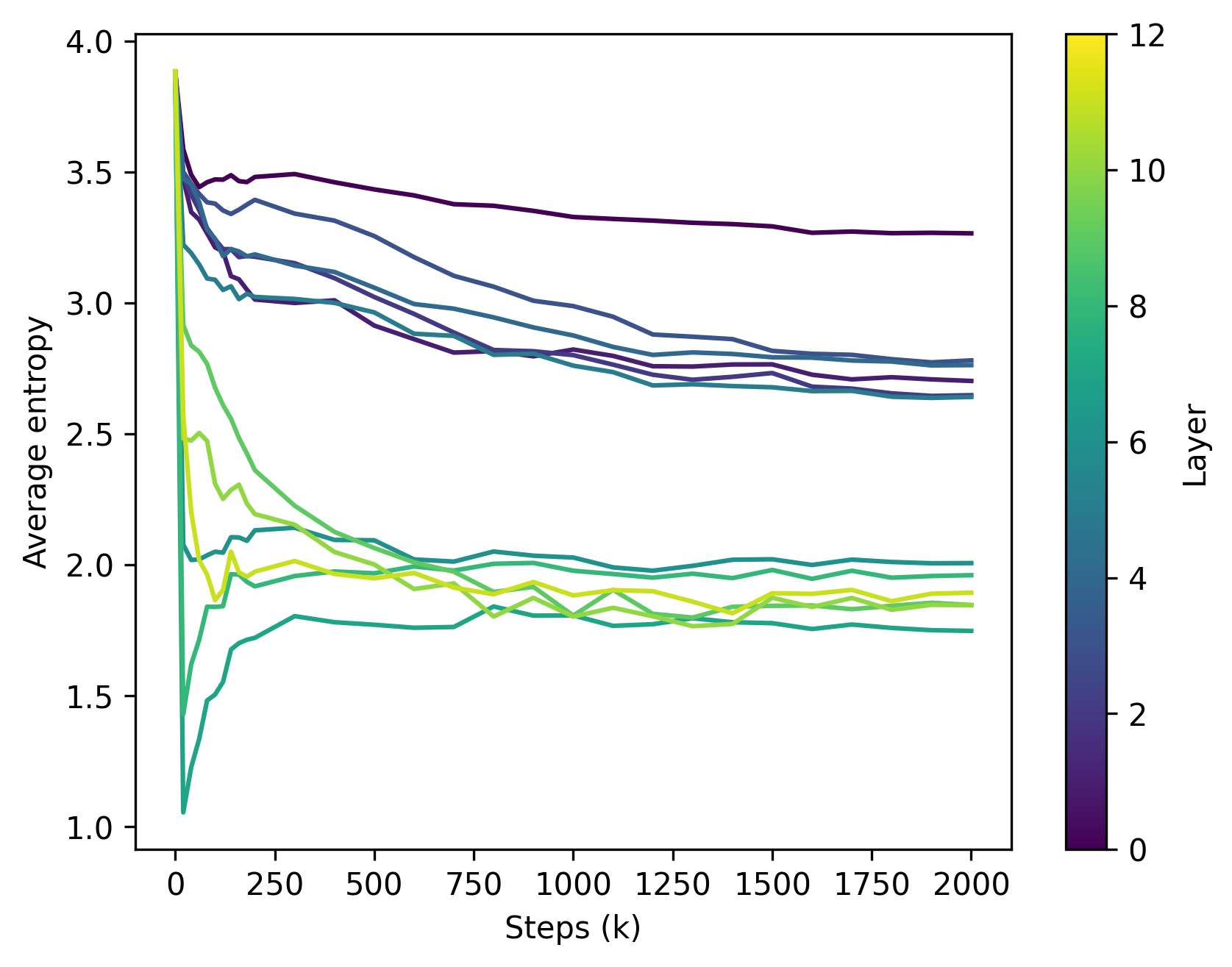}
    \caption{Average entropy of the probability distributions corresponding to self-attention rows along training. Each curve corresponds to one layer.}
    \label{fig:entropy_decay}
\end{figure}

Finally, \autoref{fig:entropy_decay} describes the evolution of the average entropy in self-attention distributions. We observe that training induces an overall decay of the entropy for all layers, with different dynamics. This corresponds to sharper self-attention distributions. It is interesting to notice that the distributions in the first layers remain sharper than the ones in the last layers.

Overall, this section shows that drift anisotropy emerges in the query and key representations during the training of MultiBERT, as self-attention distributions become sharper. The drifts of queries and keys tend to align, thus increasing the magnitude of scalar products, and the general sharpness of self-attention.

Although this section focuses on the case of token-based NLP, we believe that strong attention patterns may be required when training Transformers across all modalities, potentially generating distortions in query and key distributions that account for the final observed anisotropy of the models. However, we could not extend experiments to other modalities due to the lack of released intermediate checkpoints, to the best of our knowledge.

\section{Discussion}
\label{sec:discussion}

In this work, we argue that the nature of data distributions is not solely responsible for the anisotropy observed in most hidden representations of Transformers-based models across modalities. As \autoref{sec:empirical} shows, untrained Transformers layers display a tendency towards anisotropy. Biased inputs tend to increase the variance of the attention scores and thus facilitate the emergence of sharp patterns in the self-attention mechanisms. We also show in \autoref{sec:qk} that along training, query and key distributions drift in parallel directions, which increases anisotropy in the inner representations of the Transformer layers, while allowing sharper attention patterns. As discussed in \citet{puccetti-etal-2022-outlier}, outlier dimensions in Transformers are also involved in the emergence of strong attention patterns.

\paragraph{Consistency of the SVD} In \autoref{sec:qk}, we use an SVD on the \textit{union} of $Q_s$ and $K_s$ for visualization purposes (see \autoref{fig:proj_qk_heads} and \autoref{fig:QK_dir}). It may be argued that this approach favors the emergence of a discriminative singular direction, that helps distinguish between keys and queries, thus supporting the findings in a less convincing way. To address this concern, we display alternative projections in \autoref{sec:other_projs}, where we compute the SVD on $Q_s$ or $K_s$ only, and then project all representations using this SVD. Our observations show that our findings are consistent for these alternative projections.

\paragraph{Harmfulness of anisotropy} Even though anisotropy has not been shown to be an issue in language modeling, previous works have advocated that removing anisotropy in output representations leads to better sense disambiguation abilities \citep{bihani-rayz-2021-low, bis-etal-2021-much}. Isotropic models could also improve cross-lingual alignment in multilingual language models \citep{hämmerl2023exploring}. Nevertheless, concurrent works have suggested that anisotropy may not hurt the quality of the representations \citep{ait-saada-nadif-2023-anisotropy, rudman2023stable}. We argue that anisotropy in the Transformer architecture may actually help models by allowing sharp attention patterns, but we also believe that our work can pave the way for new isotropic architectures that can easily use sharp attention patterns.

\section*{Conclusion}
In this paper, we investigated the anisotropy problem through the lens of the drift effect, and made several contributions to the understanding of this phenomenon. We demonstrated that anisotropy can be observed in language models with character-aware architectures, extended our observations to Transformers trained on other modalities, and studied anisotropy in untrained Transformers layers. We finally explored the training dynamics of the query and key distributions, and found that they drift along a shared direction hence maximizing $QK^T$ scalar products in absolute value, allowing stronger attention patterns as a result.

We conclude that anisotropy almost systematically affects Transformers on all modalities, in a way that is not always correlated with the drift of the representations. We also provide empirical evidence that anisotropy appears as an inherent property of latent distributions used in the self-attention mechanism when modeling sharp attention patterns. We hypothesize that a revision of the self-attention operation could help reduce anisotropy by facilitating the emergence of sharp attention softmax distributions without distorting the geometry of the hidden representations.

\section*{Limitations}
As mentioned in the Discussion section, we acknowledge that \autoref{sec:empirical} does not take into account the training dynamics, and only exposes some properties of the Transformer layer at initialization. We also notice that the Spearman correlation test used in \autoref{fig:pval_vs_cos} may not be well-suited for such noisy observations, as the high p-value of the GPT-2 model shows. We provide a similar graph based on the Pearson correlation in \Cref{app:pearson}.

Moreover, we are aware that our approach is not theoretically rigorous in some aspects. For instance, we don't prove that sharp self-attention patterns \textit{cannot} emerge without anisotropy in keys and queries representations. In other words, this article is focusing on exposing and \textit{correlating} factors that explain anisotropy, but we do not demonstrate theoretical properties that would help identify the \textit{causes} of anisotropy. Nevertheless, we believe that our work can pave the way for such theoretical exploration in the future.

\section*{Ethics Statement}
To the best of our knowledge, our work does not raise any ethical concern. However, as noted in \citet{freq-based-dist}, we believe that distortions in the embedding space may be related to bias in the training data, whether it is inherent to the structure of the modality (e.g. the Zipfian distribution of words), or due to human factors (e.g. geographical considerations).

\section*{Acknowledgements}
This work was funded by the last authors' chair in the PRAIRIE institute funded by the French national agency ANR as part of the ``Investissements d'avenir'' programme under the reference ANR-19-P3IA-0001. This work was granted access to the HPC resources of IDRIS under the allocation 2023-AD011013680R1 made by GENCI.

We would like to thank Roman Castagné for useful discussions that led to focusing on observing the effect of anisotropy in the self-attention process.

\bibliography{anthology,custom}
\bibliographystyle{acl_natbib}

\appendix

\section{Pearson correlation of the drift norm and anisotropy}
\label{app:pearson}

\begin{figure}[H]
    \centering
    \includegraphics[width=\linewidth]{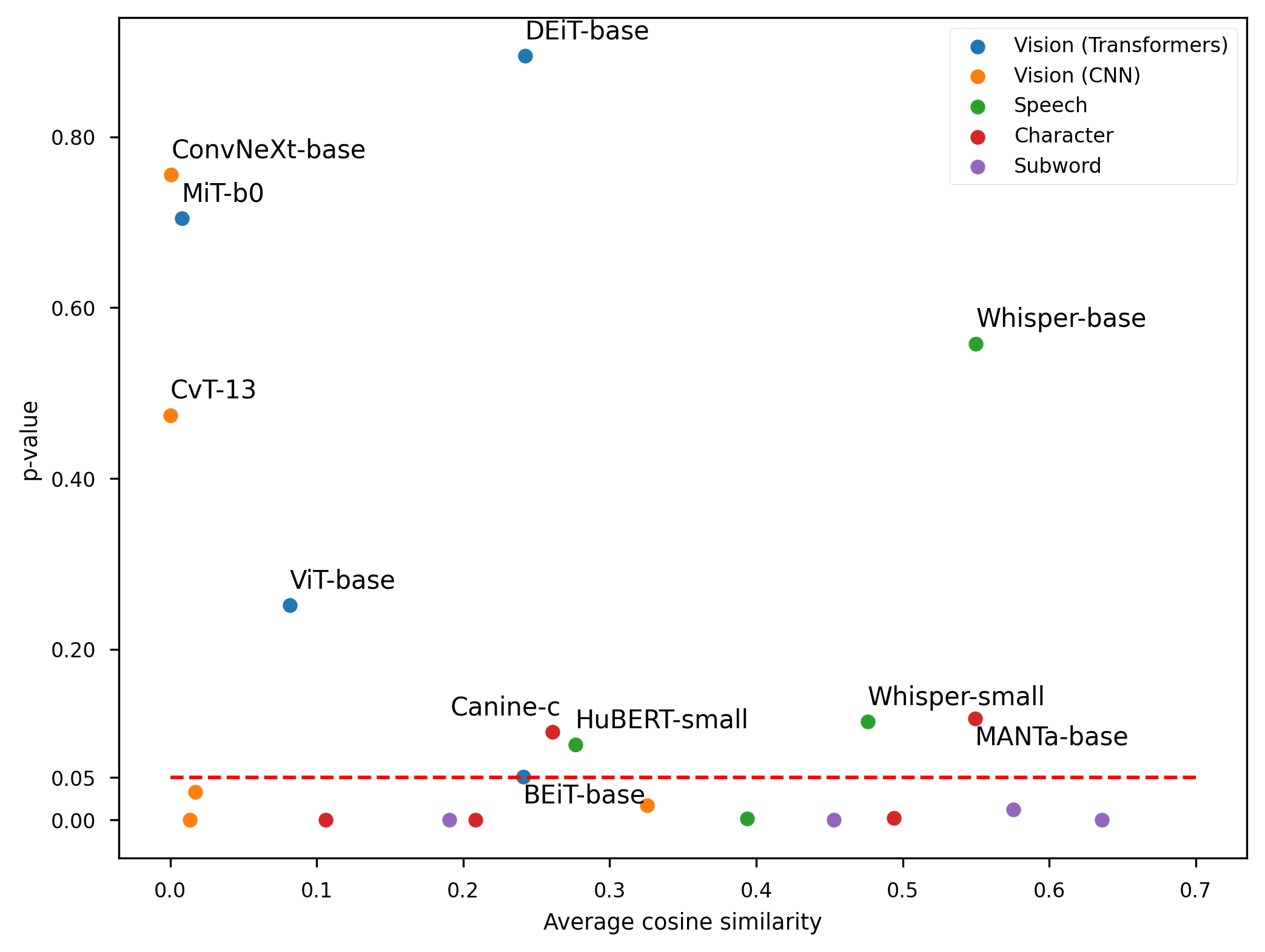}
    \caption{p-value of the Pearson correlation test between the norm of the average representation and the cosine-similarity averaged over all layers, across modalities. Models above the red dotted line are not significantly affected by the drift effect.}
    \label{fig:pval_vs_cos_pearson}
\end{figure}

The Pearson test measures a linear correlation between random variables, while the Spearman test measures a monotonic correlation. As there is no specific argument in favor of a linear relationship between the measured distributions (average cosine-similarity and norm of the average representation), we decided to use the Spearman correlation test in order to take into account more complex relation patterns.

Nevertheless, this metric is based on the rank of each observation, and is thus not robust to fluctuations due to sample variance, specifically when working with such small samples. This is reflected by the discrepancy between Pearson and Spearman p-values for some models (e.g. GPT-2).

\section{Cosine-similarity and anisotropy}
\begin{figure}[h]
    \centering
    \includegraphics[width=\linewidth]{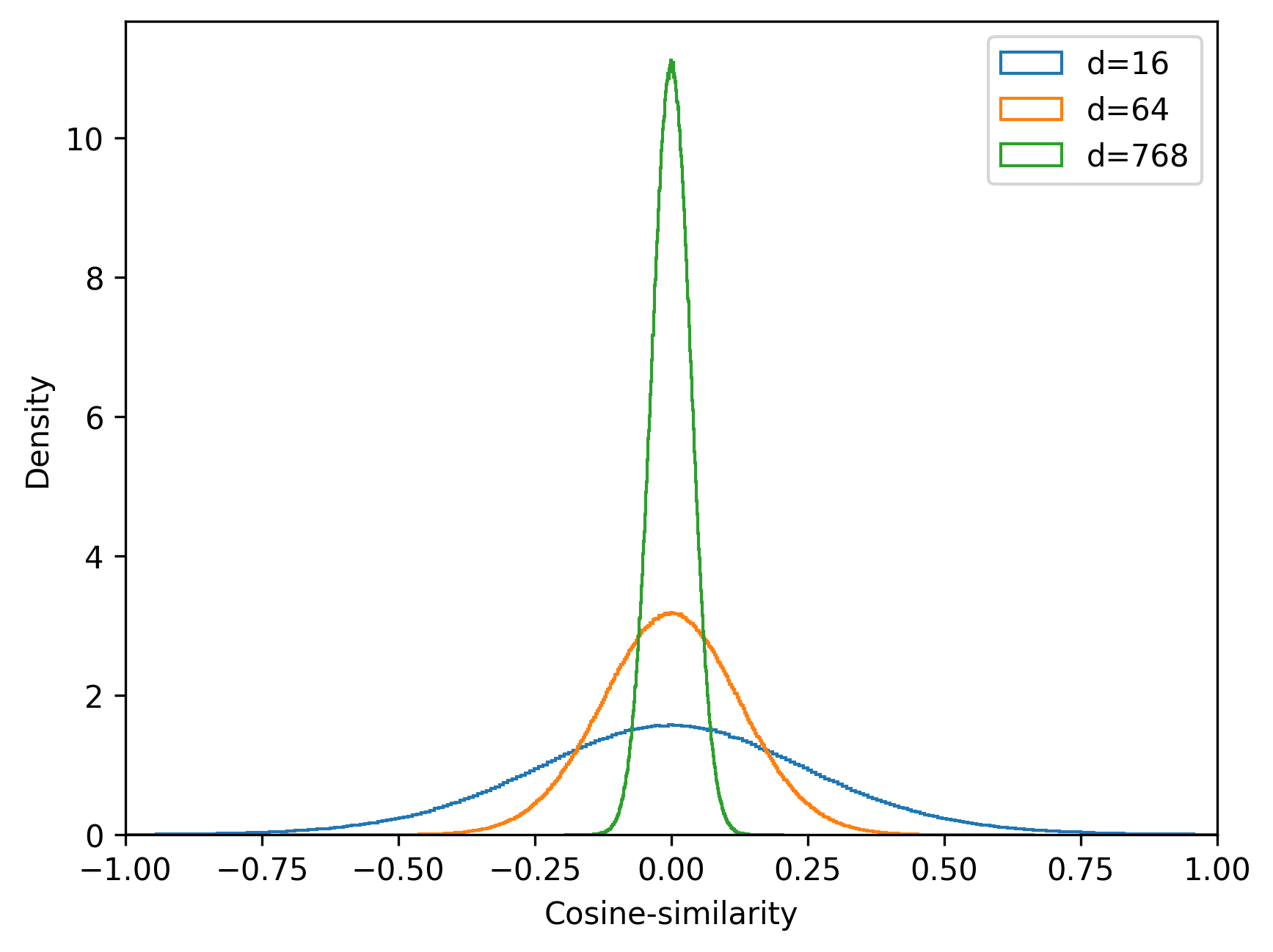}
    \caption{Density function of cosine-similarity for a normal distribution as the dimension increases.}
    \label{fig:cosine_v_density}
\end{figure}
\begin{figure}[h]
    \centering
    \includegraphics[width=\linewidth]{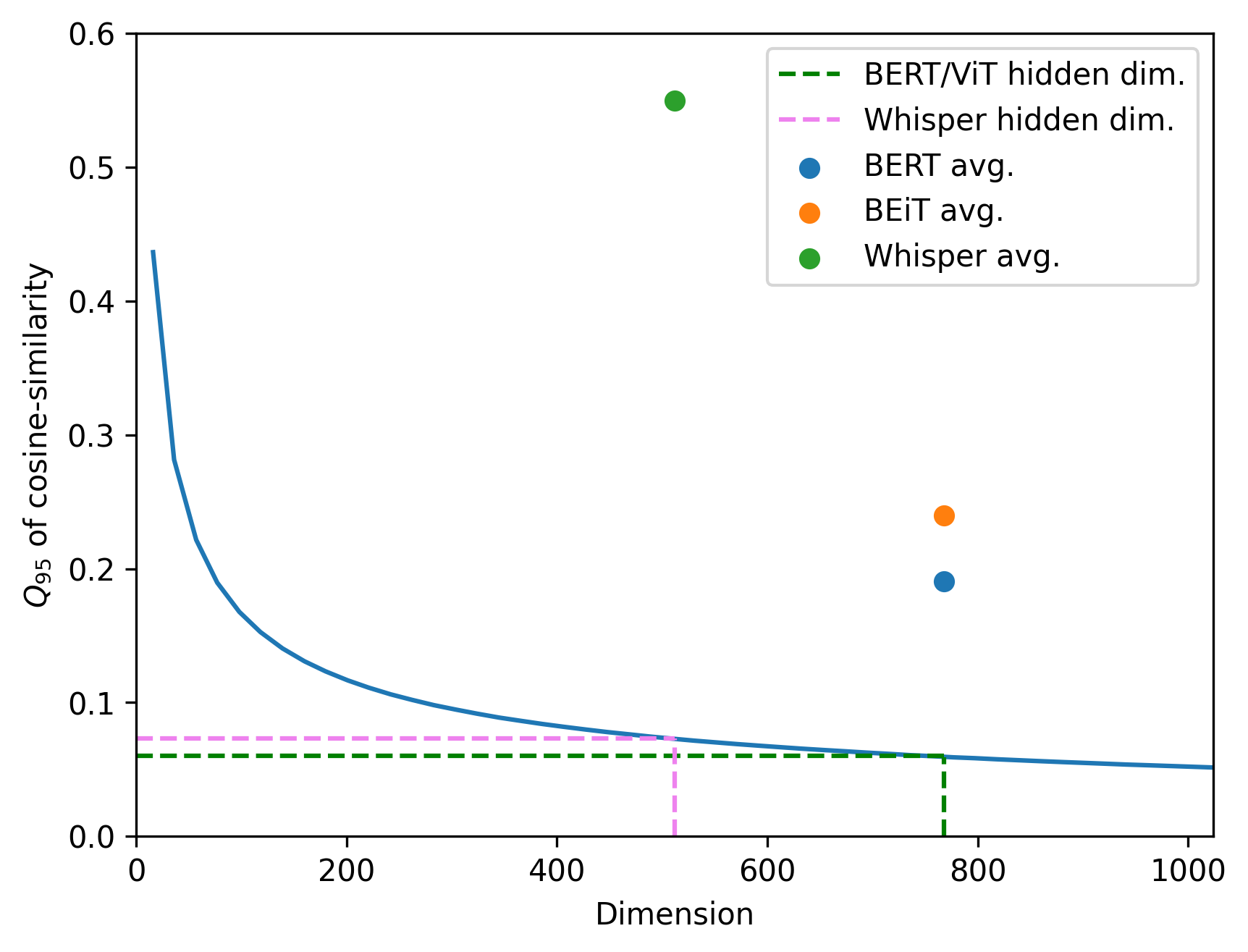}
    \caption{95th quartile of the cosine-similarity distribution on a normal distribution as the dimension increases. We add points for the average cosine-similarity level of Transformers models for several modalities.}
    \label{fig:q95}
\end{figure}

It can be argued that describing anisotropy as the observation of "high" cosine-similarity values is not a convincing definition. This section aims at showing which ranges of cosine-similarity values are characteristic of anisotropic distributions. 
In \autoref{fig:cosine_v_density}, we show the density function of the cosine-similarity values obtained when drawing pairs of samples from isotropic normal distributions in $\mathbb{R}^d$ as $d$ increases. 

For smaller dimensions ($d=16$), we see that the range of cosine-similarity values that are attained between isotropic distributions is relatively broad compared to the possible spectrum ($[-1, 1]$). As $d$ increases, the support of the observed distributions seems to become smaller, due to the curse of dimensionality.

We analyze this effect more in-depth in \autoref{fig:q95}, where we plot the 95th quantile of the cosine-similarity distribution in the isotropic scenario. We also add values for the layer-wise average cosine-similarity levels of typical models in several modalities for comparison. We can clearly observe that the levels of cosine-similarity observed in the representations of Transformers-based models are significantly unlikely to be observed in between samples drawn in isotropic normal distributions.

Nevertheless, as we go towards higher dimensional spaces for bigger models (e.g. Llama-65B from \citet{touvron2023llama} has 8192 hidden dimensions), we believe that it may be relevant to introduce isotropy metrics that are grounded to isotropic cosine-similarity distributions. We leave this question for future works.

\section{Other projections for $Q_s$ and $K_s$}
\label{sec:other_projs}
\begin{figure*}[h]
    \centering
    \begin{subfigure}[b]{0.24\linewidth}
         \includegraphics[width=\linewidth]{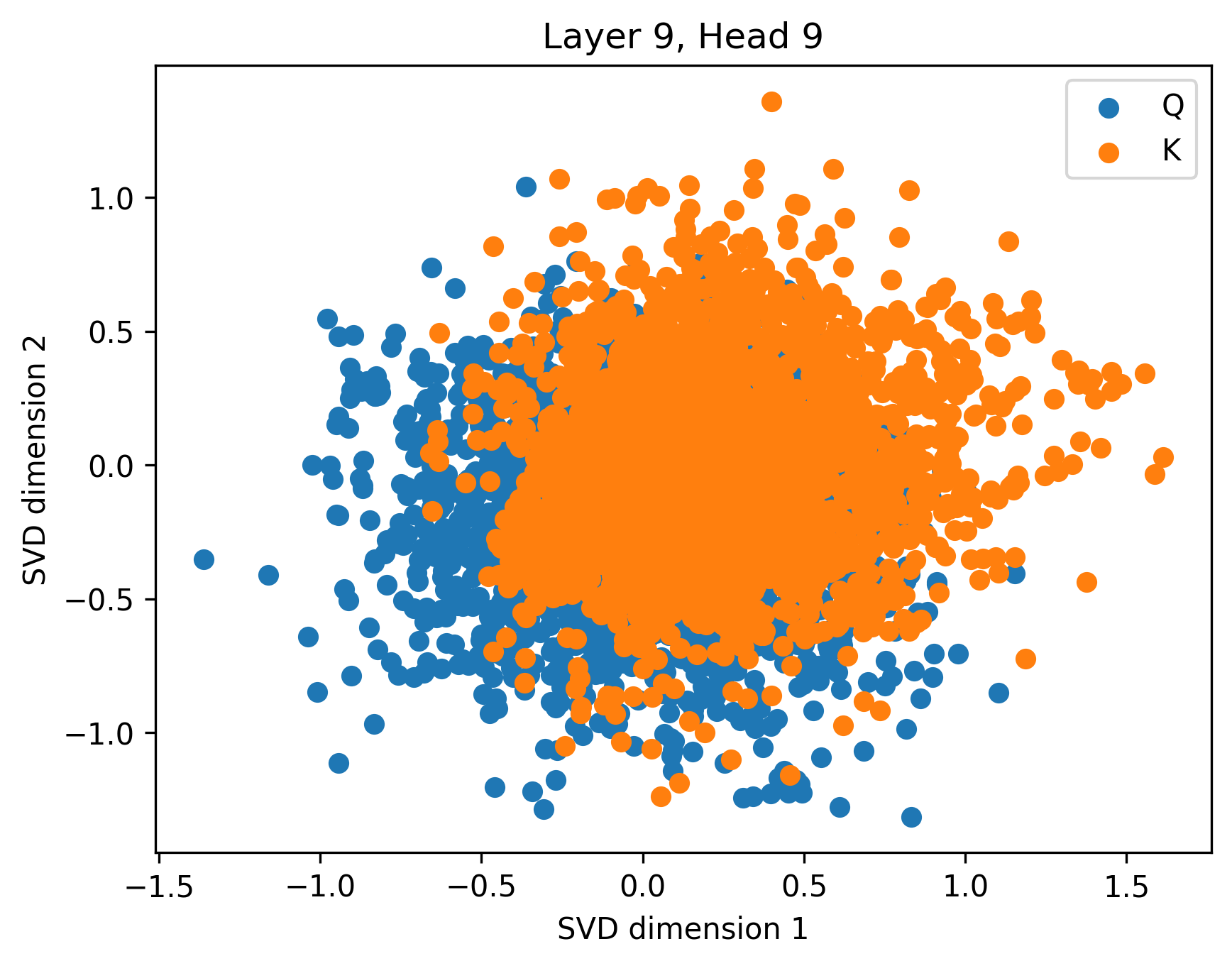}
         \caption{Step 0}
         \label{fig:dist_qk_s0_K}
    \end{subfigure}
    \begin{subfigure}[b]{0.24\linewidth}
         \includegraphics[width=\linewidth]{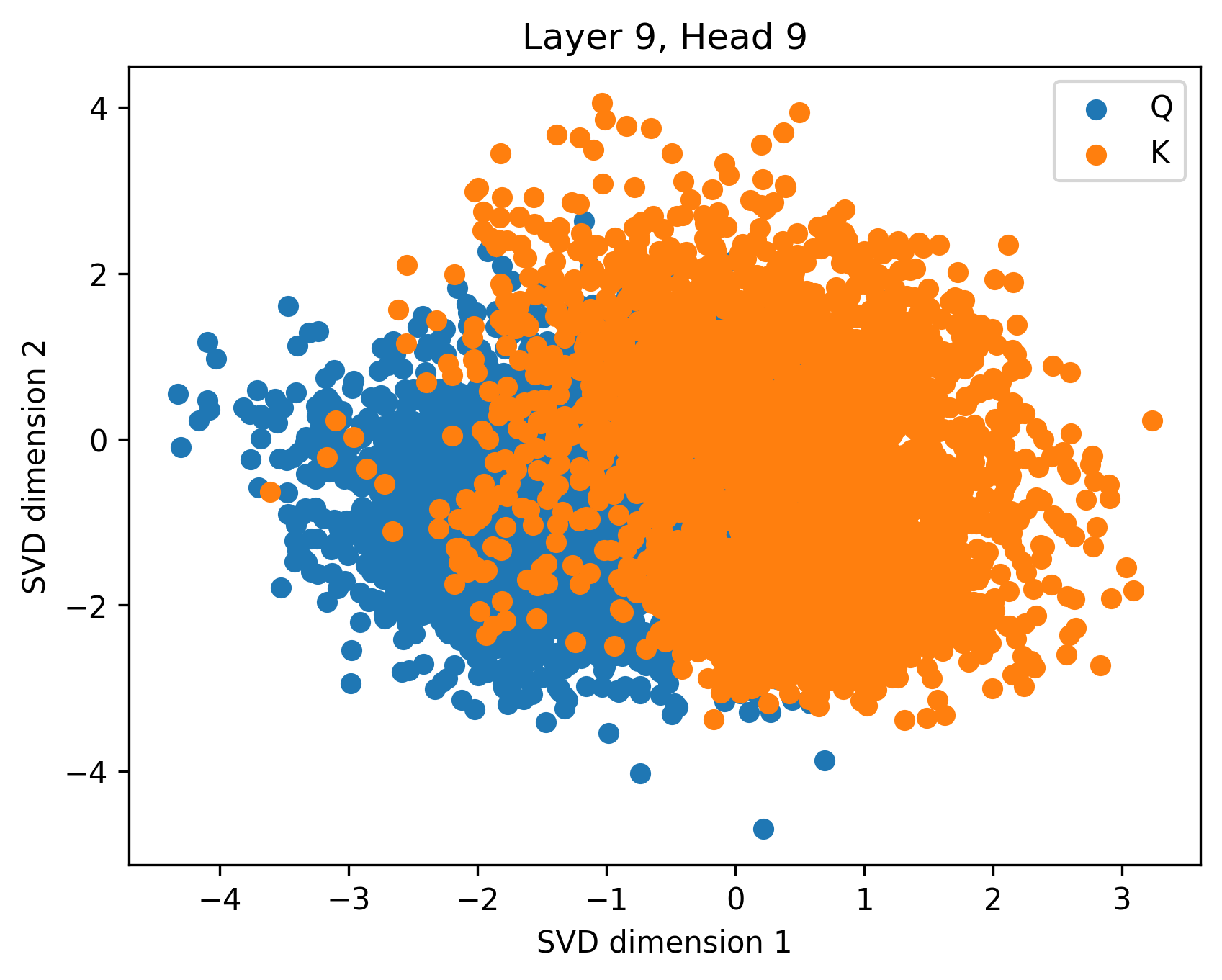}
         \caption{Step 40k}
         \label{fig:dist_qk_s40_K}
    \end{subfigure}
    \begin{subfigure}[b]{0.24\linewidth}
         \includegraphics[width=\linewidth]{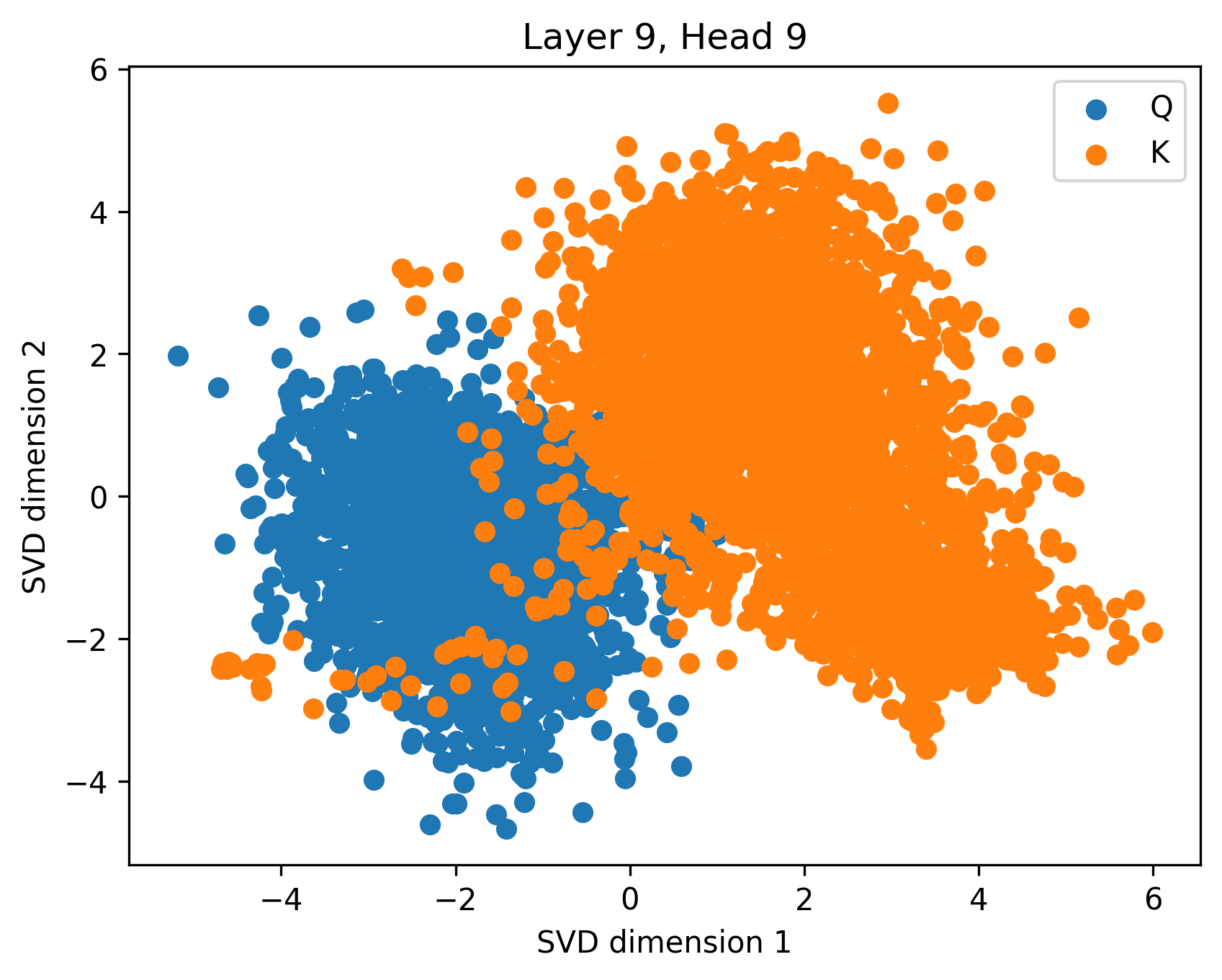}
         \caption{Step 200k}
         \label{fig:dist_qk_s200_K}
    \end{subfigure}
    \begin{subfigure}[b]{0.24\linewidth}
         \includegraphics[width=\linewidth]{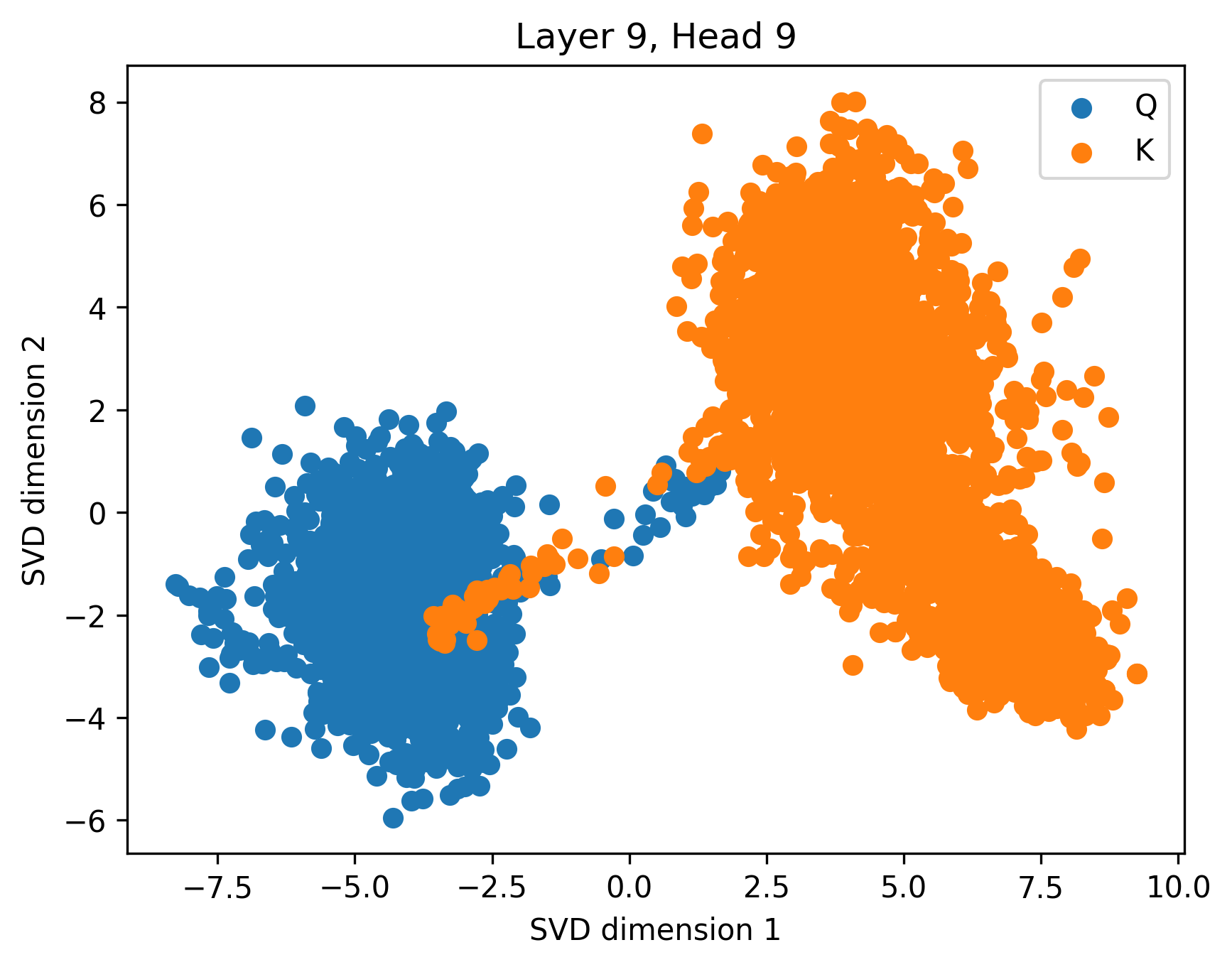}
         \caption{Step 2M (final)}
         \label{fig:dist_qk_s2M_K}
    \end{subfigure}
    \caption{Evolution of $Q_s$ and $K_s$ distributions along training. Vectors are projected using the SVD computed on $K_s$.}
    \label{fig:proj_qk_heads_K}
\end{figure*}

\begin{figure*}[h]
    \centering
    \begin{subfigure}[b]{0.24\linewidth}
         \includegraphics[width=\linewidth]{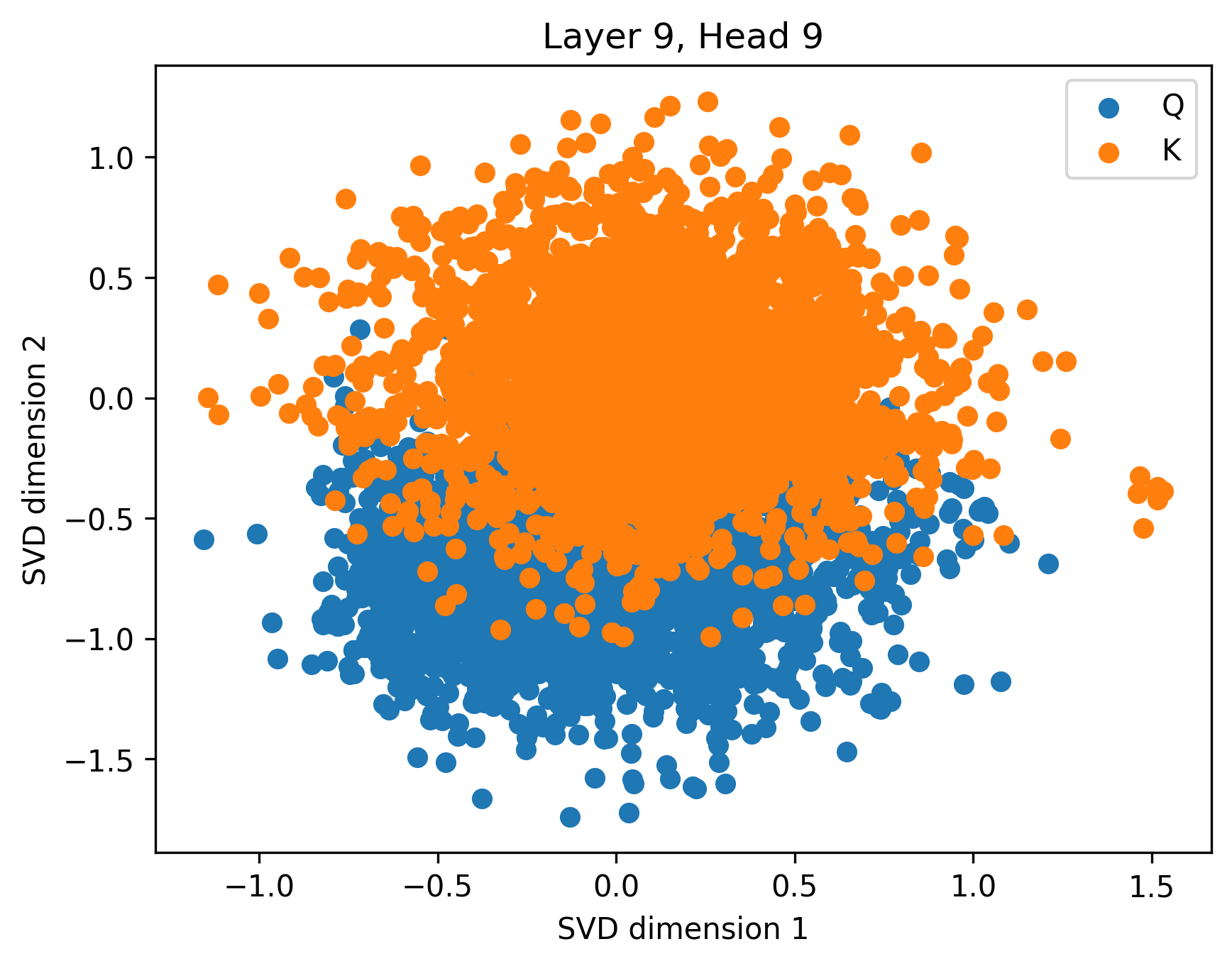}
         \caption{Step 0}
         \label{fig:dist_qk_s0_Q}
    \end{subfigure}
    \begin{subfigure}[b]{0.24\linewidth}
         \includegraphics[width=\linewidth]{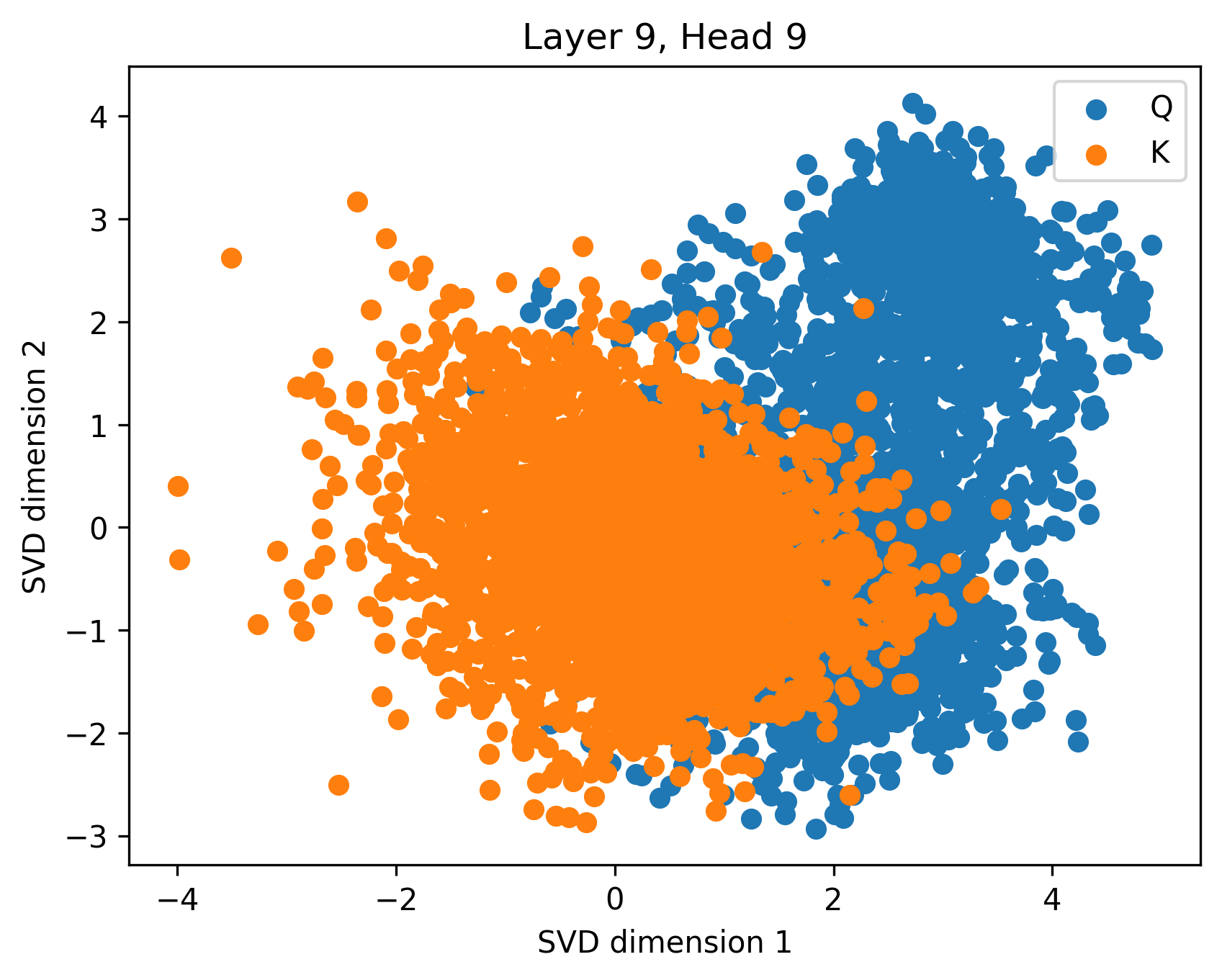}
         \caption{Step 40k}
         \label{fig:dist_qk_s40_Q}
    \end{subfigure}
    \begin{subfigure}[b]{0.24\linewidth}
         \includegraphics[width=\linewidth]{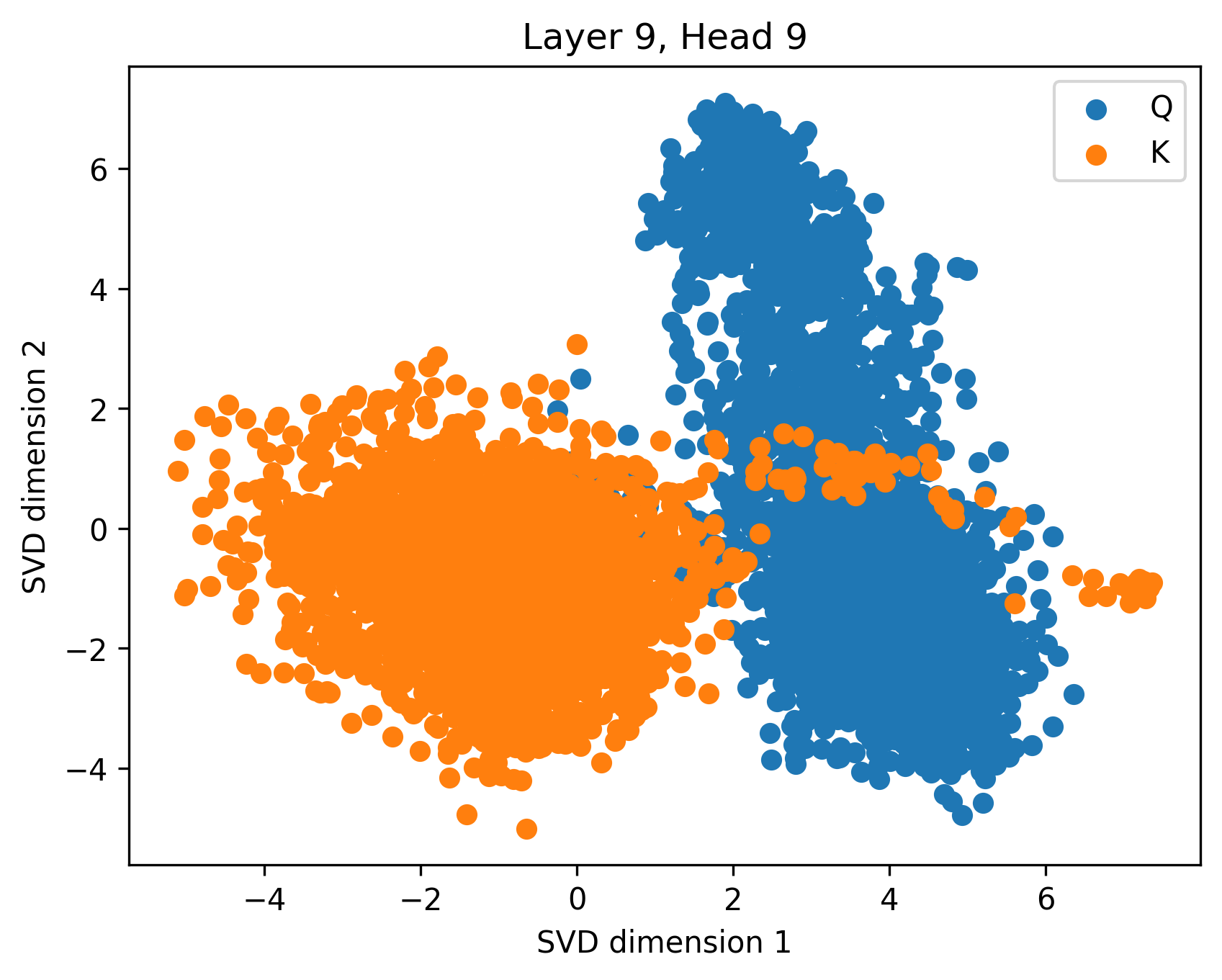}
         \caption{Step 200k}
         \label{fig:dist_qk_s200_Q}
    \end{subfigure}
    \begin{subfigure}[b]{0.24\linewidth}
         \includegraphics[width=\linewidth]{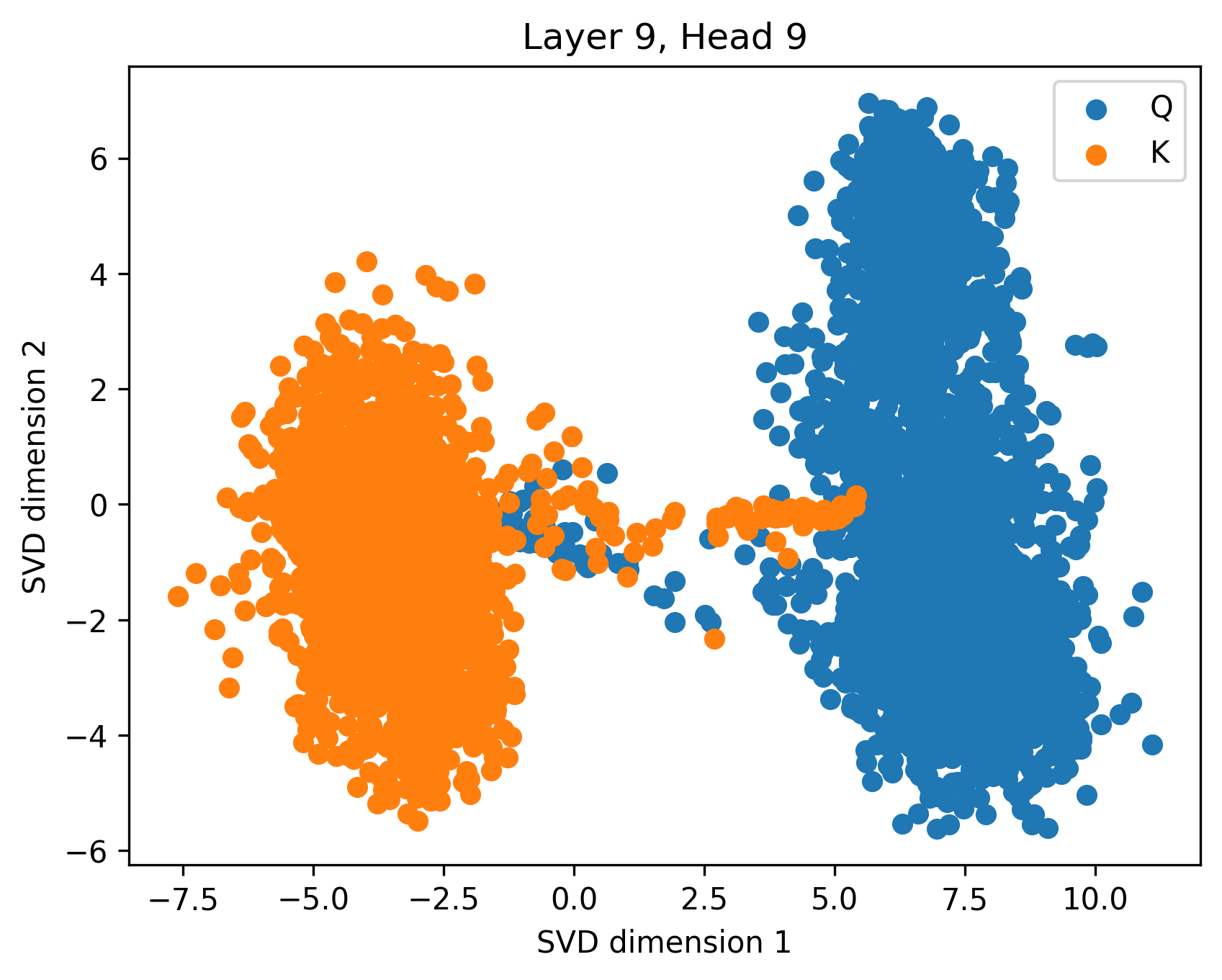}
         \caption{Step 2M (final)}
         \label{fig:dist_qk_s2M_Q}
    \end{subfigure}
    \caption{Evolution of $Q_s$ and $K_s$ distributions along training. Vectors are projected using the SVD computed on $Q_s$.}
    \label{fig:proj_qk_heads_Q}
\end{figure*}

\begin{figure}[h!]
    \centering
    \begin{subfigure}[b]{0.48\columnwidth}
         \includegraphics[width=\linewidth]{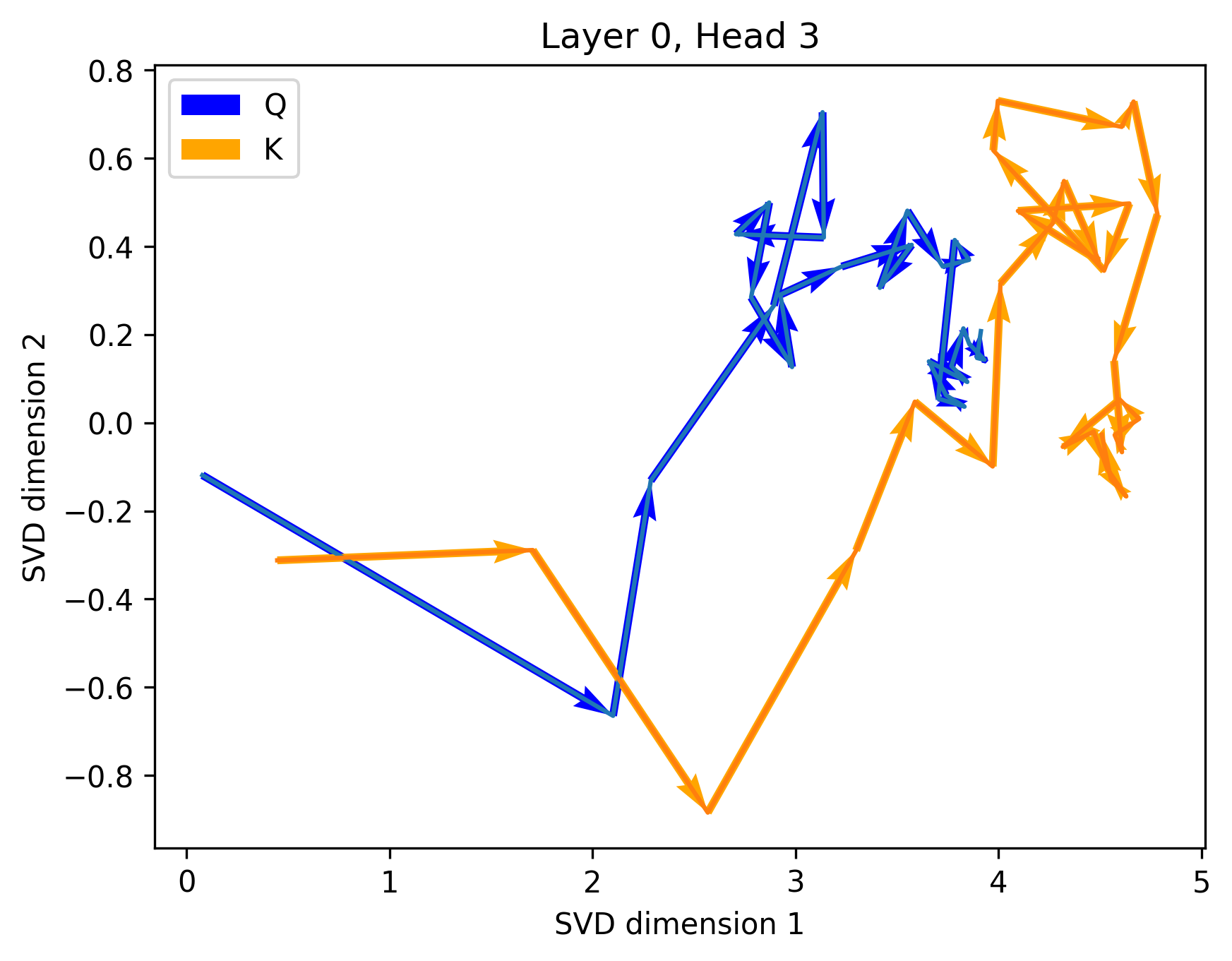}
         \caption{Similar}
         \label{fig:QK_simdir_K}
    \end{subfigure}
    \begin{subfigure}[b]{0.48\columnwidth}
         \includegraphics[width=\linewidth]{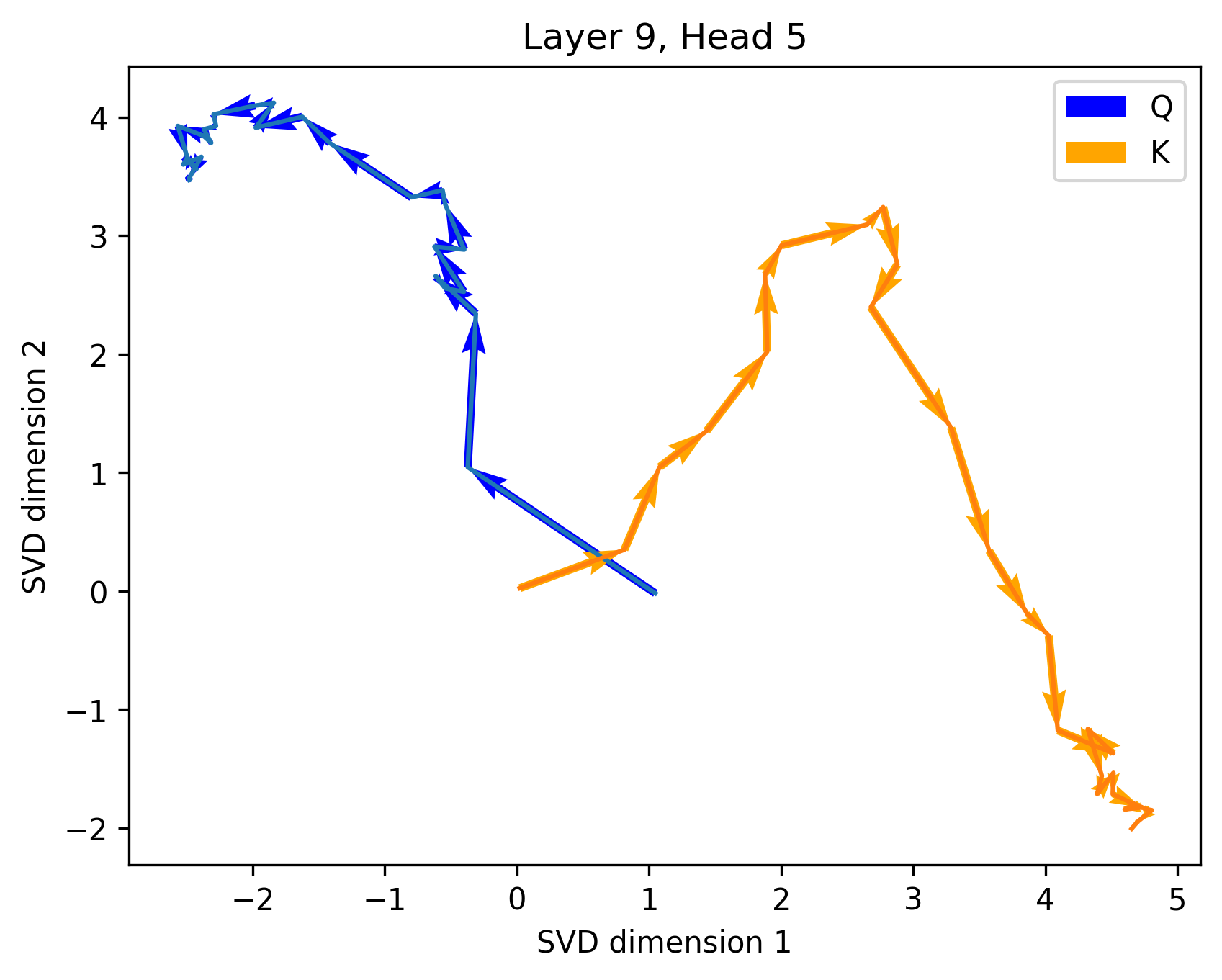}
         \caption{Opposite}
         \label{fig:QK_diffdir_K}
    \end{subfigure}
    \caption{Evolution of $\bar{Q_s}$ and $\bar{K_s}$ along training for two different heads in the network, projected via the SVD of $K_s$.}
    \label{fig:QK_dir_K}
\end{figure}

\begin{figure}[h!]
    \centering
    \begin{subfigure}[b]{0.48\columnwidth}
         \includegraphics[width=\linewidth]{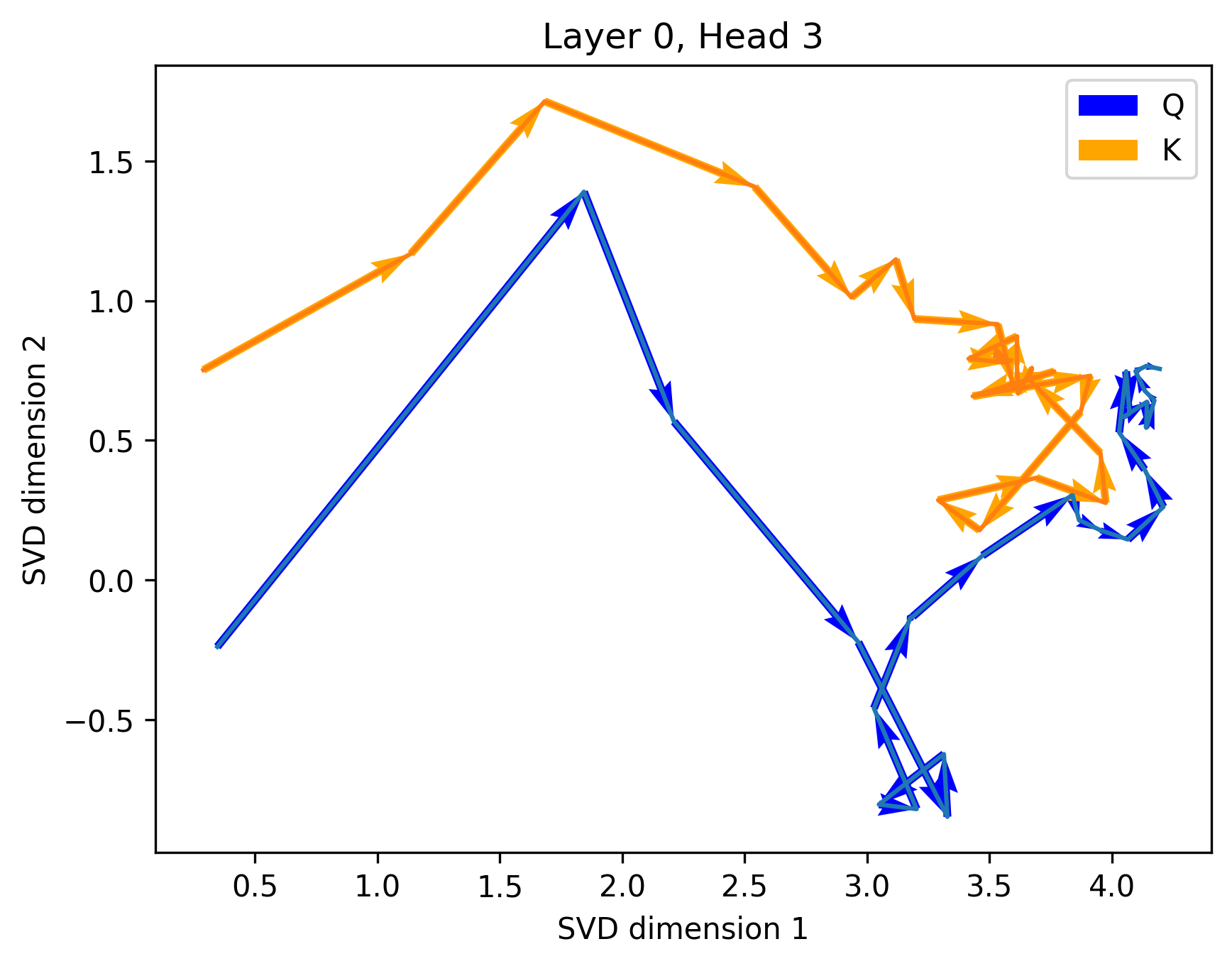}
         \caption{Similar}
         \label{fig:QK_simdir_Q}
    \end{subfigure}
    \begin{subfigure}[b]{0.48\columnwidth}
         \includegraphics[width=\linewidth]{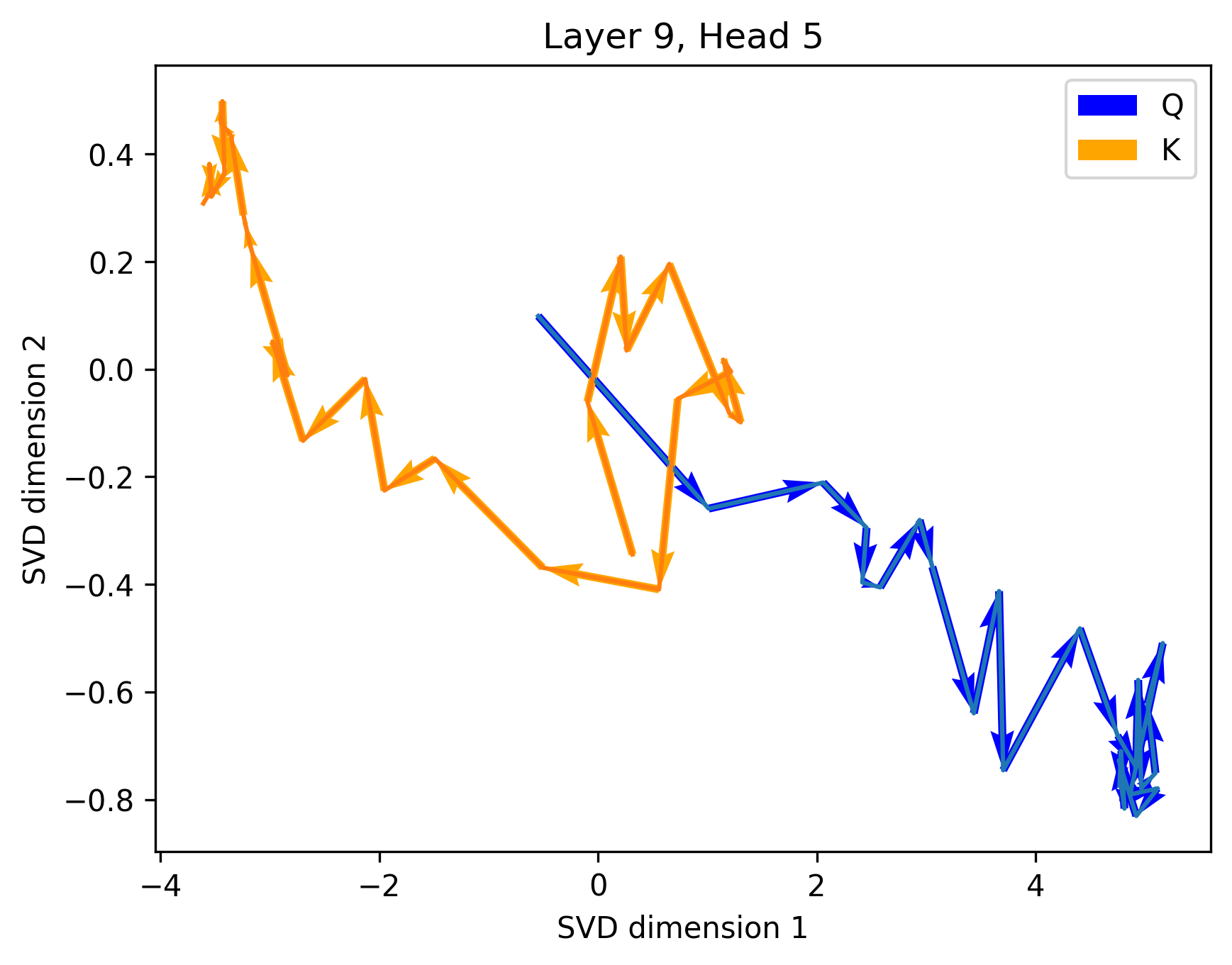}
         \caption{Opposite}
         \label{fig:QK_diffdir_Q}
    \end{subfigure}
    \caption{Evolution of $\bar{Q_s}$ and $\bar{K_s}$ along training for two different heads in the network, projected via the SVD of $Q_s$.}
    \label{fig:QK_dir_Q}
\end{figure}

As mentioned in the Discussion (\autoref{sec:discussion}), we reproduce visualizations from \autoref{sec:qk} using different projection choices. Namely, we compute the SVD on $K_s$ only in \autoref{fig:proj_qk_heads_K} and \autoref{fig:QK_dir_K}, and on $Q_s$ only in \autoref{fig:proj_qk_heads_Q} and \autoref{fig:QK_dir_Q}.

The plots show that not only does the distribution used for the SVD drifts away from the origin along training, but also that the other distribution drifts away from the origin in an opposite direction. In other words, the singular components of each distribution are also relevant to describe the drift of the other distribution. Hence, \autoref{fig:proj_qk_heads_K} and \autoref{fig:proj_qk_heads_Q} support our conclusion that the drift directions of keys and queries are aligned during training.

\section{Stability across MultiBERT seeds}
\begin{figure*}[h]
    \centering
    \begin{subfigure}[b]{0.24\linewidth}
         \includegraphics[width=\linewidth]{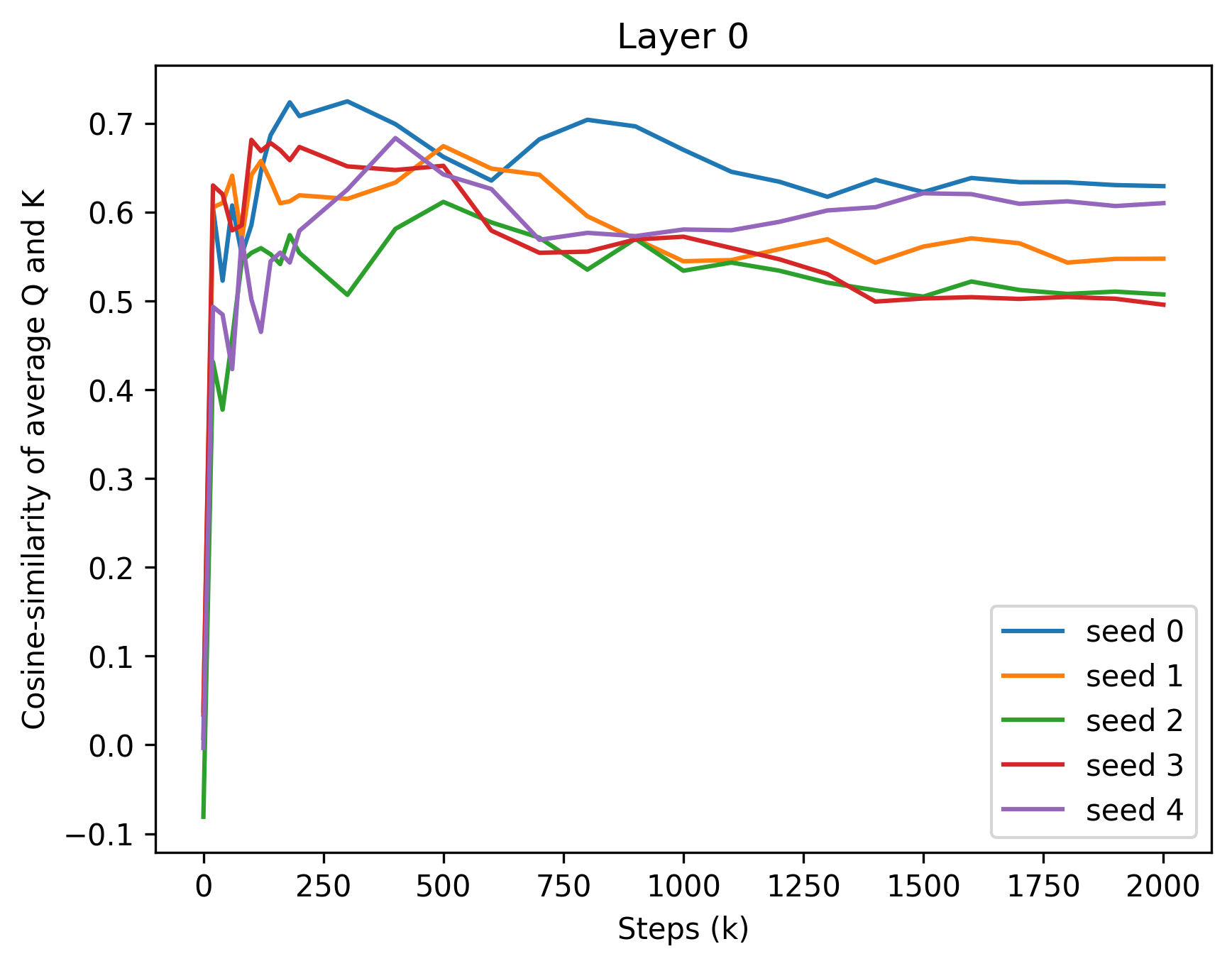}
         \caption{Layer 0}
         \label{fig:seeds_l0}
    \end{subfigure}
    \begin{subfigure}[b]{0.24\linewidth}
         \includegraphics[width=\linewidth]{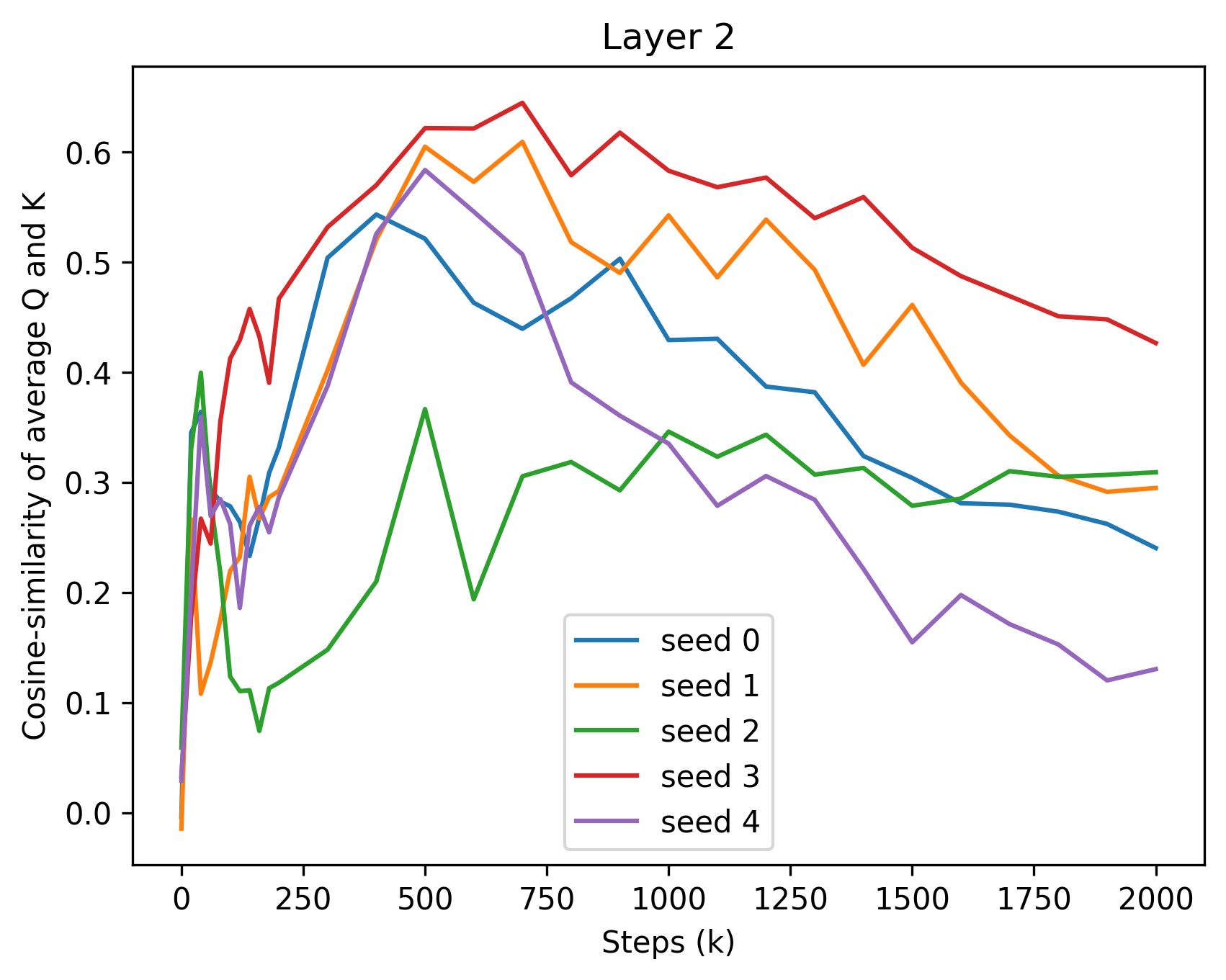}
         \caption{Layer 2}
         \label{fig:seeds_l2}
    \end{subfigure}
    \begin{subfigure}[b]{0.24\linewidth}
         \includegraphics[width=\linewidth]{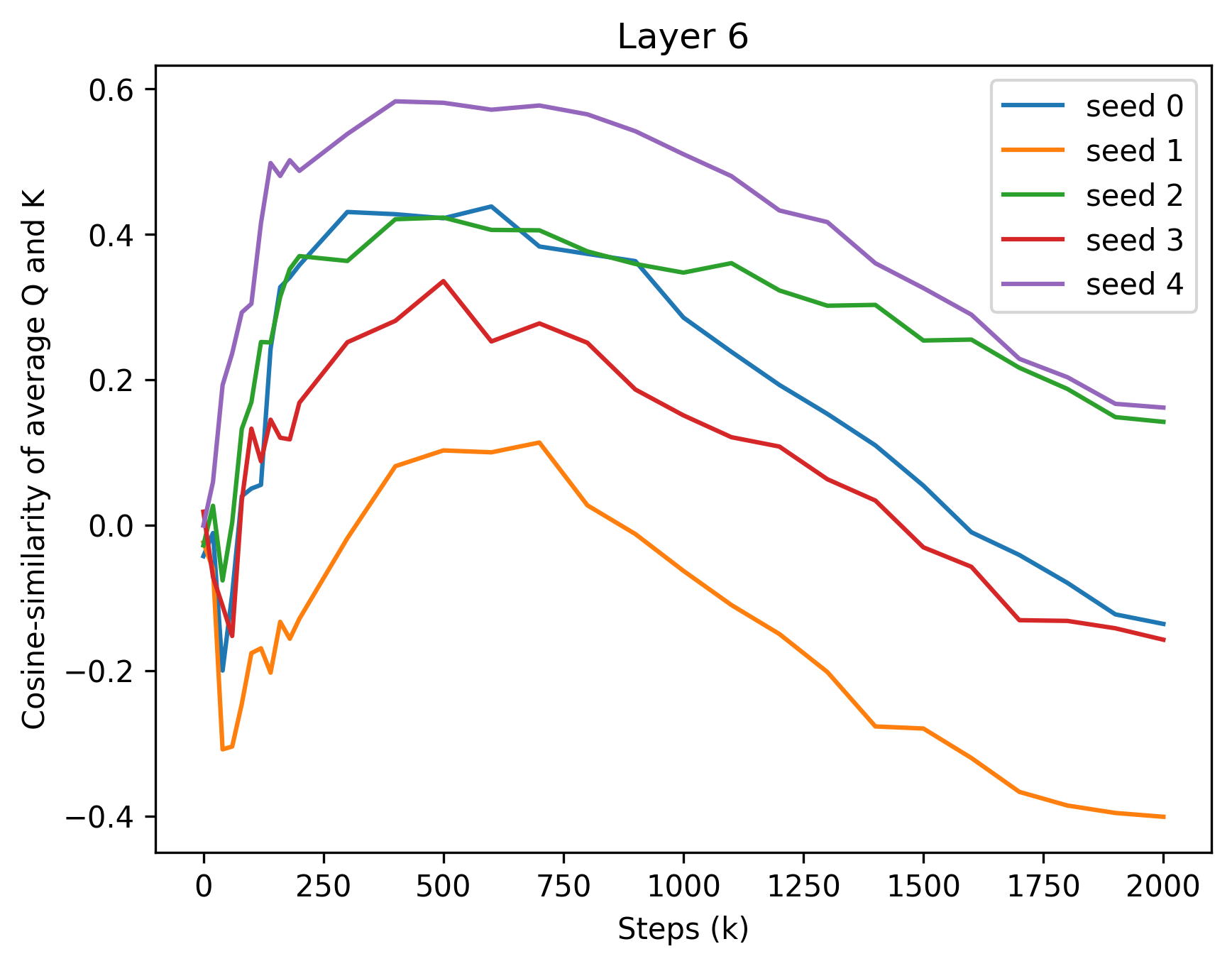}
         \caption{Layer 6}
         \label{fig:seeds_l6}
    \end{subfigure}
    \begin{subfigure}[b]{0.24\linewidth}
         \includegraphics[width=\linewidth]{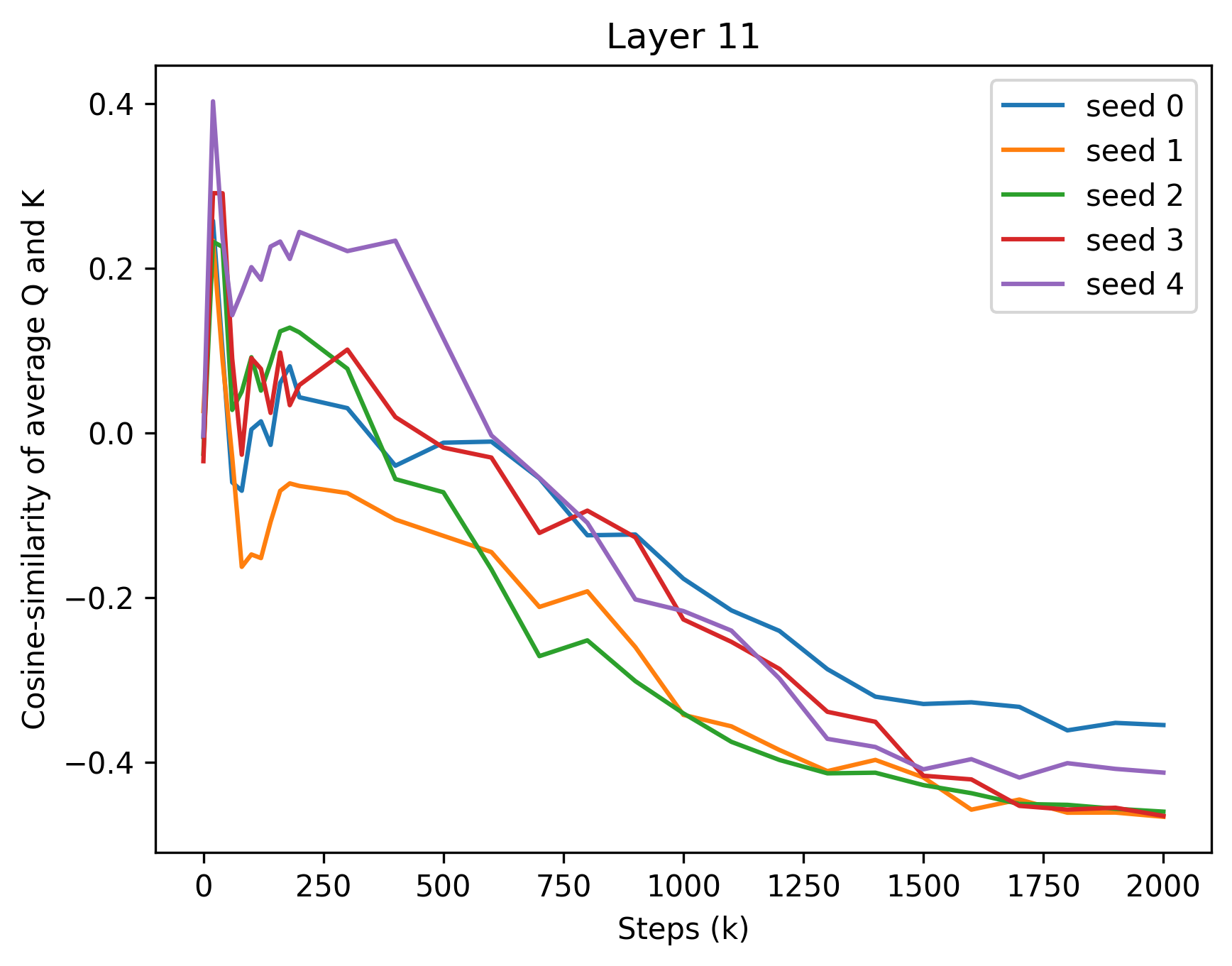}
         \caption{Layer 11}
         \label{fig:seeds_l11}
    \end{subfigure}
    \caption{Evolution of cosine-similarity between $\bar{Q_s}$ and $\bar{K_s}$ along training for various initialization seeds. Representations are concatenated across heads, and each color represents one seed of the MultiBERT models. We observe similar trends across seeds.}
    \label{fig:seeds_qk}
\end{figure*}

\end{document}